\documentclass[sigconf, nonacm = TRUE]{acmart}

\AtBeginDocument{%
  \providecommand\BibTeX{{%
    \normalfont B\kern-0.5em{\scshape i\kern-0.25em b}\kern-0.8em\TeX}}}

\newcommand{\vect}[1]{\boldsymbol{#1}}

\usepackage{optidef}

\begin{document}
\title{The Pursuit of Algorithmic Fairness: On "Correcting" Algorithmic Unfairness in a Child Welfare Reunification Success Classifier}


\author{Jordan Purdy}
\authornote{Both authors contributed equally to this research.}
\email{jordan.e.purdy@dhsoha.state.or.us}
\author{Brian Glass}
\authornotemark[1]
\email{brian.d.glass@dhsoha.state.or.us}
\affiliation{%
  \institution{Oregon Department of Human Services (ODHS), Office of Reporting, Research, Analytics, and Implementation (ORRAI)}
  \streetaddress{500 Summer St. NE}
  \city{Salem}
  \state{Oregon}
  \postcode{97301}
}

\begin{abstract}
  The algorithmic fairness of predictive analytic tools in the public sector has increasingly become a topic of rigorous exploration.  While instruments pertaining to criminal recidivism and academic admissions, for example, have garnered much attention, the predictive instruments of Child Welfare jurisdictions have received considerably less attention.  This is in part because comparatively few such instruments exist and because even fewer have been scrutinized through the lens of algorithmic fairness.  In this work, we seek to address both of these gaps.  To this end, a novel classification algorithm for predicting reunification success within Oregon
  Child Welfare is presented, including all of the relevant details associated with building such an instrument.  The purpose of this tool is to maximize the number of stable reunifications and identify potentially unstable reunifications which may require additional resources and scrutiny. Additionally, because the algorithmic fairness of the resulting tool, if left unaltered, is unquestionably lacking, the utilized procedure for mitigating such unfairness is presented, along with the rationale behind each difficult and unavoidable choice.  This procedure, though similar to other post-processing group-specific thresholding methods, is novel in its use of a penalized optimizer and contextually requisite subsampling.  These novel methodological components yield a rich and informative empirical understanding of the trade-off continuum between fairness and accuracy.  As the developed procedure is generalizable across a variety of group-level definitions of algorithmic fairness, as well as across an arbitrary number of protected attribute levels and risk thresholds, the approach is broadly applicable both within and beyond Child Welfare.
\end{abstract}

\maketitle

\section{Introduction}
\label{Sec:Introduction}

The Child Welfare division of the  Oregon Department of Human Services operates under a mission to “strengthen, preserve, and reunify families” \cite{OCFSP2015_ORIGINAL}, as well as a mission to “adapt services and policy to eliminate discrimination and disparities in the delivery of human services.” \cite{ODHSSEF015_ORIGINAL} In order to advance both principles in practice, we present a novel methodology which identifies children in custody of the state who are candidates for stable reunification with their family, and which "corrects" algorithmic unfairness in the corresponding automated classification process.  The intended implementation of this process is in the form of a decision support tool for staff who make permanency-related decisions in Child Welfare.

The purpose of such a decision support tool is to advance the mission to safely reunify children with their families by identifying the probability of a failed reunification. Currently, of children who have been removed from home and placed in substitute care for at least 90 days, only 36\% will reunify with their family within one year. However, 83\% of reunifications remain stable for at least one year. Taken together, we seek to develop a classification procedure to ascertain the probability of a stable reunification in order to 1) maximize the amount and success rate of reunifications, and 2) identify high risk reunifications which may benefit from supportive services and resources (Table \ref{Tbl:CalibrationTable}).

The construction of the proposed equitable classifier involves two stages, resulting in the assessment of an individual child’s prospects for a stable reunification along an ordinal risk tier scoring system. First, a binary classifier is constructed using statistical machine learning. Such a classifier, as far as we are aware, is the first of its kind in Child Welfare at large. Second, because such a classifier may suffer from myriad forms of bias, we administer a post-processing fairness correction. This procedure adjusts the binary classification threshold dependent on the child’s level of their protected attribute (e.g., demographic group membership). By combining multiple binary classification thresholds, a multi-tiered ordinal scoring system can be constructed. Here, a four-score system is developed using three thresholds. 

\subsection{Making Space for Algorithmic Fairness}
Machine learning classifiers have become widespread in the private sector \cite{einav2014data} and their use is expanding in the public domain \cite{oswald2018algorithm}. There exist serious and well founded concerns over the inherent unfairness or bias in these algorithms. Unfairness may be introduced into machine learning classifiers from multiple sources: externally from the historical decision-making processes which generated the training data, or internally from statistical artifacts themselves \cite{veale2018fairness}. 

We propose a post-processing fairness correction technique which seeks to address unfairness of the final automated classifier, regardless of the initial sources of the unfairness.  This procedure, though similar to other post-processing group-specific thresholding methods, is novel in its unexpected use of a penalized optimizer, its contextually necessary use of subsampling, and its generalizability across any of nine group-level definitions of algorithmic fairness.  
Given that the procedure can also accommodate an arbitrary number of protected attribute levels and an arbitrary number of risk thresholds\footnote{Assuming the sample size is sufficiently large.}, the approach is broadly applicable both within and beyond Child Welfare.

This proactive approach to mitigating a lack of algorithmic fairness represents a substantial departure from the current norm among the growing number of Child Welfare jurisdictions\footnote{We are aware of efforts, at varying stages of completion, in  Allegheny County, PA, California, Douglas County, CO, Oregon, and New York.} developing and deploying predictive risk tools.  Excluding  Oregon\footnote{Information surrounding ORRAI's various predictive analytic tools and corresponding algorithmic fairness procedure can be found on the office's public-facing website: \url{https://www.oregon.gov/DHS/ORRAI/Pages/index.aspx}.  More specifically, a high-level overview of the Oregon
Child Welfare Screening Predictive Analytics Tool, as well as the technical details of the corresponding algorithmic fairness procedure, discussed and demonstrated on a freely available data set, are provided in documents available on the website \cite{ScreeningToolDevelopment2019_ORIGINAL, PGP2019_ORIGINAL}.  Note that we have since improved upon this process by utilizing a penalized optimizer to find the group-specific thresholds, which is described later in the body of this current manuscript.}, the current ``standard'' is either to ignore algorithmic fairness altogether\footnote{Either directly or indirectly through common misunderstandings of what constitutes algorithmic fairness.  See Appendix \ref{App:MisguidedConception} for more details.} or to perform an assessment-only audit.  In fact, the only application of a fairness correction procedure to a predictive analytic tool in Child Welfare, apart from the tools developed by ORRAI, is academic and illustrative in nature \cite{coston2020counterfactual}, similar to the number of academic papers pertaining to predictive policing and criminal recidivism \cite[e.g.][]{EFNSV2018, CCDRSS2019} that can be linked to \cite{AL2016}.  This de facto state of affairs in Child Welfare is undoubtedly attributable, in no small part, to the uncertainty surrounding how ``best'' to proceed.
  
To address this uncertainty, we provide the rationale behind each critical decision embedded in the procedural methodology.  We hope such transparency serves as a ``map'' for other jurisdictions to follow and, where appropriate, deviate from accordingly.  In the case of the reunification algorithm, critical decisions were necessarily informed and influenced by the stakeholders comprising the algorithm's work group\footnote{The work group for the reunification tool included representatives from the following groups/positions, listed in alphabetical order: business analysts, Child Protective Services (CPS) supervisors, CPS workers, child safety manager, Child Welfare (CW) alcohol and drug specialist, CW district managers, CW field leadership, CW leadership, current and former foster youth, current and former foster parents, data coordinator, Indian Child Welfare Act representative from Tribal Unit, lead inter-agency researcher, Mentoring Assisting Promoting Success (MAPS) worker, Morison Child and Family Services, Office of Equity and Multicultural Services, Office of Information Services, ORRAI reporting analysts, Oregon Department of Justice, paralegals, permanency consultant, permanency program manager, permanency supervisors, permanency workers, program managers, program systems support, reunification manager, safety consultant, supervisor, and teen supervisor.}, in accordance with their knowledge of and desire for Child Welfare within the state of Oregon.  Ultimately, the presented procedure can and surely will be improved upon over time, but we hope that it serves as a new ``standard'' for those Child Welfare jurisdictions seeking to proactively address algorithmic fairness.

\subsection{Structure of Paper}
In Section \ref{Sec:PredReunSuccess}, details of the algorithm, including its development and application, are provided.  In Section \ref{Sec:AlgFairnessLens}, the identified protected attribute and chosen definition of algorithmic fairness are described.  The methodology for fairness "correcting" is presented in Section \ref{Sec:FairnessCorrectionProcedure} and applied to the reunification algorithm in Section \ref{Sec:ResultsFairnessCorrectionProcedure}.  The paper concludes in Section \ref{Sec:FinalThoughts} with a discussion of the business-case utility of the developed tool, post-fairness-correction, as well as potential opportunities for improving the utilized fairness correction methodology.  Throughout these sections, extraneous details important for replicating the work and understanding subjective decisions are provided in identified appendices. In order to encourage adoption of the technique presented here, the R code for the fairness correction procedure is posted on GitHub\footnote{\url{https://github.com/JPurdy-ORRAI/ORRAI_AlgFairnessCorrectionDemo}}, along with an illustrative R script applying the procedure to the Adult Data Set from the UCI Machine Learning Repository\footnote{\url{https://archive.ics.uci.edu/ml/datasets/Adult}}.

\section{Predicting Reunification Success}
\label{Sec:PredReunSuccess}

\subsection{Data Sources, Variables, and Outcome}
\label{SubSec:DataVarOutcome}
The primary data source was an administrative data set queried from the state of Oregon's Child Welfare data system. The data system is compliant to the U.S. Children’s Bureau’s requirements for a Statewide Automated Child Welfare Information System (SACWIS) \cite{FedCodeSACWIS}.

The general unit of observation was a child’s transition from one placement setting to another (i.e., a child-transition pair). The machine learning training set consisted of only child-reunification pairs, defined as the child-transition pairs representing a return of the child from a substitute care setting to a home setting with the child’s parents. These reunification observations consisted of both “trial visits” (i.e., temporary reunifications during which the child remains in state custody) and full reunifications (i.e., discharge from state custody, with or without ongoing government provided support services). The data set queried from this data source represented child welfare administrative data from August 2011 to January 2020. 

The outcome of interest was whether a child’s potential reunification with his/her family will be stable. To quantitatively define this outcome \cite{passi2019problem}, the binary dependent variable was deemed “true” if the child remained at home for a period of one year, and “false” if the child returned to substitute care for at least 14 days during the ensuing one-year period after returning home. Most children who have experienced at least one reunification event have experienced multiple ($25^{th} percentile$ = 1, $M$ = 2.7, $75^{th} percentile$ = 3). After stochastically unduplicating by child (i.e., choosing one reunification event per child), the overall outcome prevalence for a reunification failure was 17\%.

The independent variables (i.e., the machine learning feature set), were constructed from data elements which would have been temporally available on the day before a child-reunification pair occurred. In this way, temporal leakage was prevented by ensuring the machine learning classifier could not “peek” at information about the child’s new setting nor about the child’s future administrative data pattern. Features were only constructed using timestamped data elements with consistent data entry availability throughout the life of the SACWIS system. The features were constructed using information regarding prior: service placements in substitute care, home-based government provided child welfare service involvement, reports of abuse/neglect, and Child Protective Services (CPS) investigations. The features convey information related not only to the child, but also the child’s parents and the perpetrator listed on the child’s most recent CPS investigation. The complete list of features is available in Table \ref{Tbl:FeatureList} in Appendix \ref{App:FeaturesForClassifier}.

\subsection{From Machine Learning Classifier to Decision Support Tool}
\label{SubSec:ModelToTool}
Careful procedural protocols are required to avoid introducing artifacts via the modeling procedure itself. These “modeling pitfalls” include the selective label problem, repeated observation or temporal leakage, data volume as predictor, shrinking outcome windows, and inappropriate performance metrics. Each are discussed in Appendix \ref{App:ModelBestPractices}.

A machine learning classifier is trained to predict the outcome with the available child-reunification-level feature set. The classifier was trained using gradient-boosted decision trees via the XGBoost algorithm \cite{chen2015xgboost}. Because SACWIS system data are not conducive to exploring or understanding the causal mechanisms underlying reunification failure or success, we used a decision tree classifier to maximize predictive performance through the leveraging of complex interactions between variables \cite{breiman2001random}. The classifier training procedure involves an exhaustive resampling approach to ensure different events involving the same child do not appear in both the training and test data sets. The procedure yields an independent test-set-predicted-probability of reunification stability for each child-reunification observation in the complete set (see: Appendix \ref{AppSubSec:SelectiveLabeling})



To conform to the design and implementation specifications of Oregon's Child Welfare governance, the classifier’s output was adapted for use as a decision support tool. To accomplish this, three risk thresholds were selected to represent usable and meaningful risk tiers for practitioners, resulting in four ordinal risk score tiers\footnote{To facilitate intuitive understanding among tool users, the threshold (Prior to applying the fairness-correction procedure) separating scores of S1 and S2 from scores of S3 and S4 is the average predicted probability outputted by the algorithm.  This enables staff using the tool to quickly identify children with scores of S1 and S2 as having less than the "typical" (i.e., average) risk of a failed reunification, and children with scores of S3 and S4 as having more than the "typical" (i.e., average) risk.  The threshold separating a score of S1 from higher scores is the median of the predicted probabilities less than the average predicted probability.  This provides staff with two equal-sized "bins" in which scores of S1 convey approximately half the risk of scores of S2.  The threshold separating a score of S4 from lower scores represents the 75$^{th}$-percentile of predicted probabilities greater than the average predicted probability.  This choice simultaneously limits the proportion of child-reunification pairs receiving a score of S4 and maintains an approximate doubling of risk with each subsequent increase in risk score.}.  These thresholds are labeled low-, average-, and high-risk, while the corresponding four risk score tiers are denoted S1, S2, S3, and S4, where, for example, the low-risk threshold separates risk scores of S1 from risk scores of S2, S3, and S4.  Such coarse grained risk tier information seeks to reduce the introduction of bias via variable and uncontrollable decision thresholds which can vary between human decision makers \cite{chouldechova2018case, green2019disparate}. Finally, an extended data set of observations were constructed to consider performance generalization to child-placement observations 90 days from initial entry into substitute care. This definition mirrored the intended implementation protocol for this decision support use case.

\subsection{Predictive Performance}
\label{SubSec:PredPerf}
To assess predictive performance, Table \ref{Tbl:CalibrationTable} reports the model's predictive characteristics for child-reunification pairs as well as for children who remained in substitute care at 90 days after entry. As the classifier was trained using child-reunification pairs, the Reunification group represents a direct predictive test of the classifier. Table \ref{Tbl:CalibrationTable} indicates that of the Reunifications with a high-risk score (i.e., S4), 51\% resulted in a reunification failure (i.e., a return to substitute care). In contrast, of those with a low-risk score (i.e., S1), 7\% resulted in a failure.

The In Care group represents a potential use case of the classifier to assess children in substitute care at 90 days from entry. In this sense, the In Care group is a generalized use of the classifier, in that it represents an observation type that is not used in the training of the classifier. Note that 36\% of the children in the In Care group experience a reunification event in the upcoming year. This proportion represents the critical opportunity to provide a decision support tool to advance the mission to reunify children with their families. In particular, from Table \ref{Tbl:CalibrationTable}, it is apparent that among the 27\% of children who have been in care for 90 days and would receive a low-risk score (i.e., S1), only 40\% will experience an attempted reunification in the next year; whereas, among the 8\% of children who have been in care for 90 days and would receive a high-risk score (i.e., S4), a comparatively high 32\% will experience an attempted reunification in the next year.  The lack of clear separation in these and the other values of the \textit{\%Reunify w/in 1 Year} column of Table \ref{Tbl:CalibrationTable} suggest that the diagnosticity of the tool could greatly enhance reunification-related decisions in Oregon Child Welfare.

The area under the receiver operating characteristic curve (AUC/ROC) was 0.73. Despite this moderate AUC value, the low historical reunification rate coupled with the lack of diagnosticity of the historical reunification decisions represents a real opportunity to support human decision making in this area. The scores provided in Table \ref{Tbl:CalibrationTable} are generated using the fairness corrected version of the classifier. The fairness correction procedure is discussed below.



\begin{table}[h]
  \caption{Calibration table for risk score outcomes and proportions for two groups: 1) children who leave substitute care to reunify with their parents, and 2) children who have not yet reunified with their parents at 90 days from entry into substitute care.}
    \label{Tbl:CalibrationTable}
\begin{tabular}{c|c|c|c|c}
 \toprule
        & \multicolumn{2}{c|}{\textbf{Reunifications}} & \multicolumn{2}{c}{\textbf{In Care at 90 Days}}  \\
        & \% Given    & \% Failed        & \% Given  & \% Reunify                                \\  
Score   & Score       & Reunification    & Score    & w/in 1 Year                                \\  \midrule 
S4       & 8\%               & 51\%                       & 8\%                 & 32\%            \\
S3       & 24\%              & 25\%                       & 31\%                & 33\%            \\
S2       & 34\%              & 13\%                       & 34\%                & 36\%            \\
S1       & 34\%              & 7\%                        & 27\%                & 40\%            \\ \midrule
Overall & 100\%             & 17\%                       & 100\%               & 36\%            \\ 
    \bottomrule
\end{tabular}
\end{table}

\section{The Chosen Lens for Algorithmic Fairness}
\label{Sec:AlgFairnessLens}
This section establishes the protected attribute and definition of algorithmic fairness agreed upon by the stakeholders comprising the reunification algorithm work group.  Maximizing the extent to which this definition of fairness is achieved across the identified levels of the protected attribute, without detrimentally impacting predictive performance of the tool, is the focus of the "correction" procedure described in Section \ref{Sec:FairnessCorrectionProcedure}.  
\subsection{The Protected Attribute}
\label{SubSec:IdentifiedPA}
A race and ethnicity-based protected attribute is utilized for the reunification algorithm. This feature consists of four levels:  Black (BL); Hispanic, Pacific Islander, or Asian (HPA); Native American or Indian Child Welfare Act (ICWA)-eligible (NV); and White (WH).  Importantly, BL and NV child-reunification pairs are approximately 47\% and 39\% more likely, on average, to experience a failed reunification than HPA child-reunification pairs, and approximately 20\% and 14\% more likely, on average, to experience the adverse event than WH child-reunification pairs.  See Appendix \ref{App:IdentfyingPA} for further details surrounding the identification and leveling of this protected attribute, as well as the specifics of the aforementioned disproportionality. 

\subsection{The Definition of Algorithmic Fairness}
\label{SubSec:DefnAlgFair}
Many definitions of algorithmic fairness for a binary classification task have been proposed in the literature.  An overview of the different categories of definitions, as well as many of the proposed definitions within each category, is provided in \cite{VR2018}.  For the reunification algorithm, the group-level category of definitions, as opposed to the individual-level or causal reasoning-based categories, was ultimately chosen.  Short summaries of these categories, along with the logic guiding our decision to focus on the category of group-level definitions, is provided in Appendix \ref{App:WhyGroupLevel}.

The particular group-level definition utilized for the reunification algorithm is Error Rate Balance, which is equivalent to Equalized Odds, as defined in \cite{HPS2016}.  The decision to use this definition, as opposed to any of the other alternative group-level definitions, reflects the collective input, knowledge, and values of the stakeholders making up the work group for the reunification algorithm.  Brief overviews of nine group-level definitions, along with our rationale for choosing Error Rate Balance, is provided in Appendix \ref{App:WhyErrorRateBalance}.  Recognizing that algorithmic fairness is not simply, or perhaps even at all, a problem of mathematical correctness, but instead a challenge around building an algorithm that supports and furthers human values \cite{N2018}, we consider the choice of definition to be  jurisdiction- and use-case-dependent.  For this reason, the fairness "correction" procedure presented below has embedded functionality enabling its broad application across a variety of group-level definitions, including all of those discussed in Appendix \ref{App:WhyErrorRateBalance}. 

\subsubsection{Contextualizing Error Rate Balance}
Error Rate Balance, at each risk score threshold, requires equality across all protected-attribute-level-specific false negative rates, and across all protected-attribute-level-specific false positive rates.  For a given child-reunification pair outcome, be it "success" or "failure", this means the probability of an incorrect prediction label must be the same across all levels of the protected attribute.  This definition therefore recognizes that the algorithm, like all algorithms, is going to make these mistakes, but it requires them to be proportionately experienced across the levels of the protected attribute.  

\subsubsection{Quantifying Error Rate Balance}
\label{SubSubSec:QuantifyingERB}
To quantify the extent to which Error Rate Balance is achieved\footnote{As all group-level definitions are technically met or unmet by the algorithm, we require a method for quantifying the extent to which the definition is achieved.  The measure presented in this manuscript is applicable with any of the other group-level definitions of algorithmic fairness discussed in Appendix \ref{App:WhyErrorRateBalance}} for a single random subsample of child-reunification pairs, we identify at each threshold the greatest disparity in either false negative rates or false positive rates across any pairing of protected attribute levels.  This value, at each threshold, is calculated such that it ranges continuously between 0 and 1, with larger values indicating greater similarity in false positive rates and false negative rates.  For example, a value of $0.5$ indicates that at least one level of the protected attribute has an error rate that is half that of some other protected attribute level at the specified threshold. In Appendix \ref{App:QuantifyingErrorRateBalance}, we exemplify in detail the calculation of this measure.  To accommodate the dependence structure of the observational units, this Error Rate Balance measure is then calculated, at each threshold, across a large number of random subsamples.  The corresponding average value for each threshold represents an audit of the extent to which Error Rate Balance is achieved.

While such an audit is discussed in detail in Section \ref{Sec:ResultsFairnessCorrectionProcedure} for both the pre- and post-fairness-corrected risk scores, the values of the audit for the "uncorrected" risk scores are provided here to demonstrate the need for a "correction" procedure with the reunificaiton algorithm.  These values are 0.68, 0.61, and 0.59 at the low-, average-, and high-risk thresholds, respectively.  This means, for example, that at the average-risk threshold one protected attribute level is only 0.61 times as likely as some other protected attribute level to experience an incorrect prediction label.  In particular, at the average-risk threshold, the greatest disparity is between the false positive rates of BL and HPA child-reunification pairs; BL child-reunification pairs who will not experience a failed reunification within 1 year of its initiation are assigned an S3 or S4 risk score with probability 0.340, whereas for HPA child-reunification pairs that probability is only 0.206\footnote{The protected-attribute-level-specific average false negative and false positive rates for all four protected attribute levels, at all three (pre-fairness-corrected) thresholds are provided in Table \ref{Tbl:PAspecificFNRsFPRs} in Appendix \ref{AppSub:PAspecificFPRsAndFNRs}}.  Mitigating these disparities in false negative and false positive rates, at all three thresholds, is the objective of the procedure presented in Section \ref{Sec:FairnessCorrectionProcedure}.

\section{Approach to Fairness "Correcting" the Reunification Risk Scores}
\label{Sec:FairnessCorrectionProcedure}
Many procedures for mitigating a lack of algorithmic fairness have been proposed in the literature.  An overview of the different classes of procedures, as well as many of the proposed procedures within each class, is provided in \cite{BHJKR2017}.  For the reunification algorithm, a protected-attribute-level-specific thresholding adjustment procedure, within the class of post-processing procedures, has been utilized.  The logic leading to this decision is provided in Appendix \ref{App:IdentifyingFairnessCorrectionProcedure}.  

In the context of the reunification algorithm, such an approach means that the amount of evidence of success\footnote{As provided by the predicted probability, or some ordered analogue, of the classifier.} required to elevate the risk score\footnote{For example, from a score of 2 to 3.} for a child-reunification pair depends on the level of the protected attribute; furthermore, the ``dial'' for determining the ``appropriate'' amount of evidence for each protected attribute level is ``tuned'' according to the specified definition of algorithmic fairness.  In \cite{LCM2017}, this definition is Statistical Parity\footnote{Technically, the p-percent rule is used as the ``tuning'' mechanism, but the p-percent rule is just a measure quantifying the extent to which Statistical Parity is achieved.}, while in \cite{HPS2016} this definition is Error Rate Balance\footnote{Technically, the procedure is discussed both for Equal Opportunity and Equalized Odds, but Error Rate Balance is equivalent to Equalized Odds.}.   

\subsubsection{Clarifying Point}
It is worth addressing here a common and natural rhetorical question to such a thresholding adjustment procedure: But is it not unfair to uphold different ``standards'' for different protected attribute levels when assigning risk scores?  Such a response is, at its core, pushing back against the notion that to prevent disparate impact requires disparate treatment, and is in fact the motivation behind the in-processing approach proposed in \cite{ZVRG2017}.  In reality, however, such approaches seeking to prevent disparate impact without disparate treatment 1) fail to optimally prevent disparate impact and 2) ultimately do enact disparate treatment ``...through hidden changes to the learning algorithm'' \cite[][pg. 16]{LCM2017}.  Hence, such a criticism is not limited to a protected-attribute-level-specific thresholding approach, but in fact applies to a broad range of fairness correction procedures.  And in response to this more general criticism, we point out that any disparate treatment is in fact in service to a core value of the Oregon Department of Human Services -- service equity.

\subsubsection{Addressing Use-Case Realities}
\label{SubSubSec:AddressingUseCaseRealities}
While the procedure developed and utilized for the reunification algorithm attains algorithmically fairer risk scores through the same mechanism (i.e., protected-attribute-level-specific threshold values) as the procedures described in \cite{HPS2016} and \cite{LCM2017}, the optimization process for identifying the values for these thresholds is notably different and, to the best of our knowledge, novel in its application.  This optimization process simultaneously accounts for the dependence structure embedded in the observational units\footnote{Children can and frequently do experience more than one reunification event}, yields an empirically derived curve of the trade-off continuum between fairness and accuracy, and accommodates any of nine specified group-level definitions of algorithmic fairness\footnote{Including, calibration, conditional use accuracy equality, equal opportunity, error rate balance, overall accuracy equality, predictive equality, predictive parity, statistical parity, and treatment equality.}.  As such, the procedure represents an important contribution to the body of literature on algorithmic fairness in the predictive risk tools of Child Welfare.

It is important to note that this procedure will almost surely not result, either in this use case or in general, in fully achieving algorithmic fairness according to Error Rate Balance.  In an effort to combat this reality, alternative processes, including that proposed in \cite{HPS2016}, incorporate a random mechanism during the assignment of risk scores.  In the context of the reunification algorithm, for some proportion of child-reunification pairs, using such a mechanism means a risk score of S1 versus S2, for example, is driven by chance.  As such randomness is not a viable option for Child Welfare stakeholders, the procedure presented in this paper instead seeks to more completely understand the functional trade-off between accuracy and fairness so that the "best" compromise between them can be identified.

\subsection{The "Correction" Procedure}
\label{SubSec:Procedure}
The critical components of the procedure for identifying the protected-attribute-level-specific threshold values are provided in this section; the full procedure, step-by-step, is provided in Appendix \ref{App:TheFairnessCorrectionProcedureStepByStep}.  In short, this procedure utilizes a penalized optimizer within a subsampling loop to obtain the corresponding threshold values.  For the reunification algorithm, the procedure is executed using a $4 \times N$ matrix, denoted by $\vect{X}$, in which the $N$ rows represent the entire set of child-reunification pairs and the columns represent, respectively, a unique child identification number (for subsampling purposes), the observed value of the binary outcome variable, the level of the protected attribute, and the predicted probability obtained from the algorithm.

For a random subsample\footnote{One reunification per child} of $\vect{X}$, denoted $\vect{X}_i$, let $\Phi_{t,i}$ denote the group-agnostic (i.e., protected-attribute-level indifferent) threshold value for subsample $i$ at threshold $t$ (e.g., average-risk threshold), identified in accordance with the business case motivating the predictive analytic tool.  
Then, a value for the tuning parameter representing the trade-off between accuracy and fairness, denoted $w$, where $0\leq w\leq 1$, is specified.  The penalized objective function to be minimized is
\begin{equation*}
f(\vect{\theta}_{t,i}) = (1-w)(1-\mbox{ERB}(\vect{\theta}_{t,i})) \;+\; w\Delta(\vect{\theta}_{t,i}, \Phi_{t,i})
\end{equation*}
where $\mbox{ERB}(\vect{\theta}_{t,i})$ is the value quantifying the extent to which Error Rate Balance is achieved for subsample $i$ at $\vect{\theta}_{t,i}$, as described in Section \ref{SubSubSec:QuantifyingERB}, and $\Delta(\vect{\theta}_{t,i}, \Phi_{t,i})$ is the proportion of risk scores that change (e.g., either from 2 to 3 or from 3 to 2) in subsample $i$ when moving from the group-agnostic threshold value, $\Phi_{t,i}$, to the group-specific (i.e., protected-attribute-level specific) threshold values specified by $\vect{\theta}_{t,i} = \bigg(\theta_{t,i}^{P1}, \theta_{t,i}^{P2},\ldots, \theta_{t,i}^{PK}\bigg)'$, where $P1,\ldots,PK$ denote, respectively, the $K$ levels of the protected attribute\footnote{For the reunification algorithm, $K=4$, with $P1 = BL$, $P2 = HPA$, $P3 = NV$, and $P4 = WH$}.  The value of $\vect{\theta}_{t,i}$ that minimizes $f(\vect{\theta}_{t,i})$ is denoted $\widehat{\vect{\theta}_{t,i}}$. 
This objective function is sequentially\footnote{If calibration is the specified definition of algorithmic fairness, then the optimization process must be restructured into a simultaneous, rather than sequential, problem.  The details of the restructuring are provided in Appendix \ref{App:SimultaneousPenalizedOptimizationCalibration}.} utilized for each threshold, with associated pragmatic constraints updating accordingly; see Appendix \ref{AppSub:UnderstandingTheLogicOfTheProcedure} for the rationale behind these constraints. 

This sequential process is then repeated across a large number, $I$, of subsamples (e.g., $I=200$).  The pre-fairness-corrected value for threshold $t$, for a specified value of $w$, denoted $\Phi_{t,w}$, is obtained by averaging across the corresponding subsample values, $\Phi_{t,i}$.   Similarly, the post-fairness-corrected values for threshold $t$, for a specified value of $w$, $\widehat{\vect{\theta}_{t,w}}$, are obtained by averaging, component-wise, across the corresponding subsample values, $\widehat{\vect{\theta}_{t,i}}$.

\subsection{Understanding the Procedure}
\label{SubSec:ProcedureLogic}
The subsampling component of this procedure ensures that no two reunifications for the same child are simultaneously used in the objective function, which addresses concerns surrounding 1) the dependence structure of the observational units, and 2) the robstness of fairness-correction procedures to training-test splits. \cite{friedler2019comparative}. The corresponding "cost" of this approach primarily comes in the form of an increase in the overall computational burden.  This expense, however, is not necessarily significant since the subsampling qualifies as embarrassingly parallel.

To recognize how the procedure yields an empirical curve of the trade-off continuum between accuracy and fairness, consider the penalized objective function utilized with each subsample. This function measures the extent of the algorithmic unfairness in subsample $i$ through the $1-ERB(\vect{\theta}_{t, i})$ term.  Since smaller values of this term correspond to ``fairer'' group-specific threshold values, the ``fairest'' threshold values are obtained by minimizing this term.  To then quantify and capture the trade-off between fairness and accuracy, this term is penalized by the corresponding decrease in accuracy incurred as a result of adopting the group-specific threshold values over the group-agnostic threshold value.  The utilized penalty term, $\Delta(\vect{\theta}_{t,i}, \Phi_{t,i})$, is not a direct measure of any specific type of predictive accuracy, but is instead a broad catch-all for all such measures.  In particular, when a risk score changes as a result of transitioning from the group-agnostic threshold value to the group-specific threshold values, all predictive performance measures are necessarily impacted\footnote{Though this impact is not necessarily for the worse, depending on the measure.}.  Hence, the greater the proportion of changing risk scores, the greater the potential accuracy-related costs.  

The gated weighting mechanism, with weight $0\leq w\leq1$, bounds how far from the group-agnostic threshold value the optimizer is willing to search for the optimal group-specific threshold values.  When $w=0$, the optimization process will yield the fairest group-specific threshold values for subsample $i$ without any regard for accuracy.  When $w=1$, the optimization process in exclusively concerned with accuracy and, consequently, the returned group-specific threshold values for subsample $i$ will be identical to the group-agnostic threshold value.  More generally, as the value of $w$ is increased over the range of values between these extremes, the pursuit of fairness becomes more heavily anchored to the accuracy achieved through the group-agnostic threshold value.  By iterating the procedure across a tuning grid of $w$-values ranging from zero to one, the functional relationship between accuracy and fairness can empirically be explored at each threshold.

Finally, the procedure is easily applied across alternative group-level definitions of algorithmic fairness.  This is achieved by replacing $ERB(\vect{\theta}_{t, i})$ with the analogous measure for the specified alternative definition.  The specifics of such calculations are addressed in Appendices \ref{AppSub:QuantifyingAltGroupLevelDefns} and \ref{App:SimultaneousPenalizedOptimizationCalibration}. 

\section{Applying and Assessing Fairness Correction Procedure}
\label{Sec:ResultsFairnessCorrectionProcedure}
\begin{figure*}[t]
    \centering
    \includegraphics[width=\linewidth, height=2.0in]{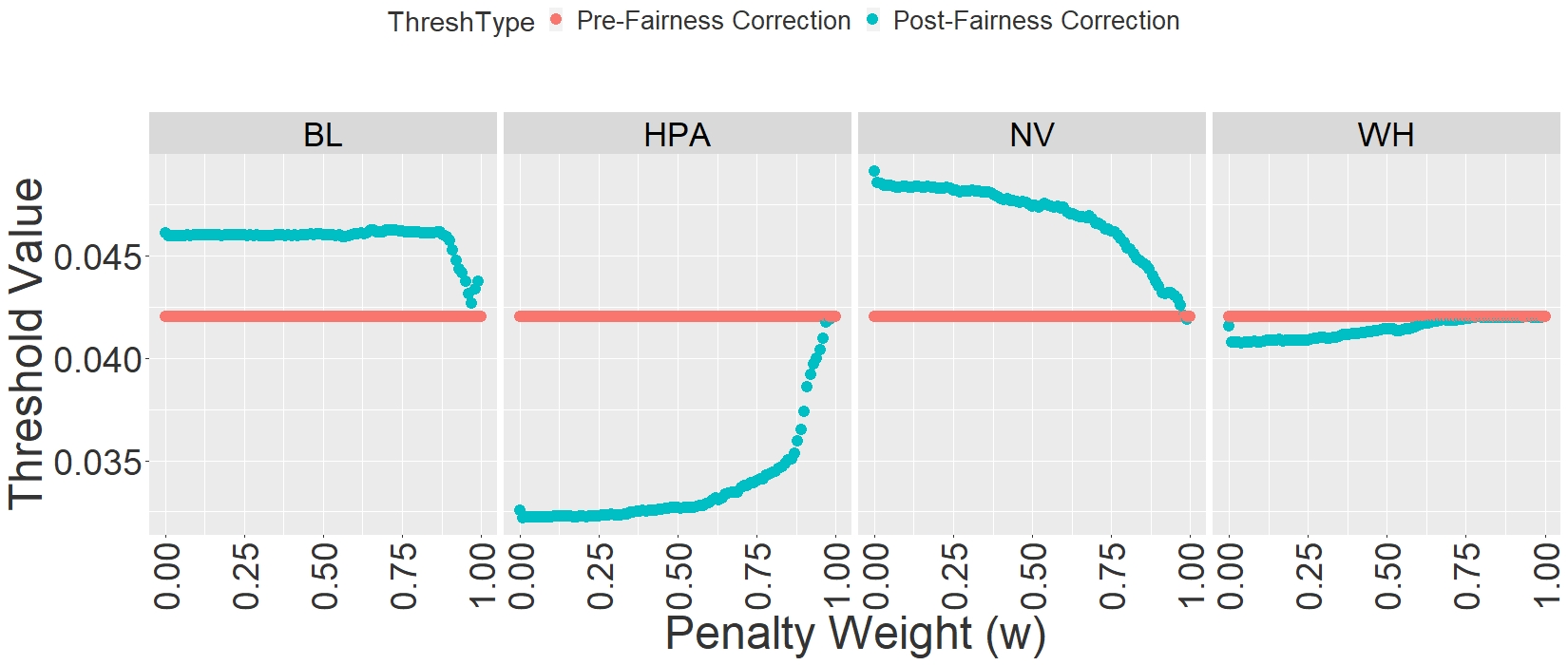}
    \caption{The pre- and post-fairness-corrected low-risk threshold values resulting from utilizing the procedure detailed in Section \ref{SubSec:Procedure} with 200 random subsamples (i.e., $I=200$) across each of 101 distinct penalty weights (i.e., $w=0, 0.01, 0.02, \ldots,0.98, 0.99, 1$).}
    \label{Figure:LowRiskThresholdsAllPenaltyWeights}
\end{figure*}

The procedure described in Section \ref{SubSec:Procedure} was applied to the reunification algorithm with 200 random subsamples (i.e., $I=200$) utilized across each of 101 distinct penalty weights (i.e., $w=0, 0.01, \ldots, 0.99, 1$).   Figure \ref{Figure:LowRiskThresholdsAllPenaltyWeights} displays the corresponding pre- and post-fairness-corrected low-risk threshold values obtained at each penalty weight; analogous results are found at both the average-risk and high-risk thresholds.   Observe in this figure that, as should necessarily be the case, the pre-fairness-corrected threshold values are the same across all four protected attribute levels and all penalty weights.  Furthermore, observe that as $w$ increases towards one, the distance between the pre- and post-fairness-corrected threshold values tends to decrease towards zero.  This is by design and demonstrates that the penalty term in the penalized objective function is operating as intended.  More specifically, as costs to accuracy become more and more ``valued'', as conveyed through the increasing value of $w$, the post-fairness-corrected threshold values are ``pulled back`` towards their corresponding ``accuracy-anchored'' pre-fairness-corrected threshold value.

\subsection{The "Best" Penalty Weight}
\label{SubSec:IdentifyingBestPenaltyWeight}
To identify the "best" penalty weight at each threshold, and thereby determine the set of post-fairness-corrected threshold values to utilize when the algorithm is implemented, the predictive performance and algorithmic fairness of these various sets of threshold values must be evaluated\footnote{The corresponding results may be optimistic since training and testing splits were not used as a result of computational limitations.  See Appendix \ref{App:NeedTrainingTestingSplitsConvo} for additional details.}.  The dependent nature of the child-reunification pairs again facilitates the need for a subsampling procedure.  While the corresponding step-by-step procedure is provided in Appendix \ref{App:IdentifyingBestPostFairnessCorrectedThresholdValuesStepByStep}, the basic idea is to apply each set of threshold values (both $\Phi_{t,w}$ and $\widehat{\vect{\theta}_{t,w}}$) to a large number, $J$, of random subsamples and compute for each an assortment of predictive performance metrics as well as the extent to which Error Rate Balance\footnote{And for completeness, the extent to which other group-level definitions of fairness are achieved.} is achieved.  The average and standard deviations of these measures, across the $J$ subsamples, is then used to identify the "best" set of post-fairness-corrected threshold values.

\subsubsection{Visualizing the Trade-Off Continuum Between Fairness and Accuracy}
\begin{figure*}[t]
    \centering
    \includegraphics[width=\linewidth, height=2.25in]{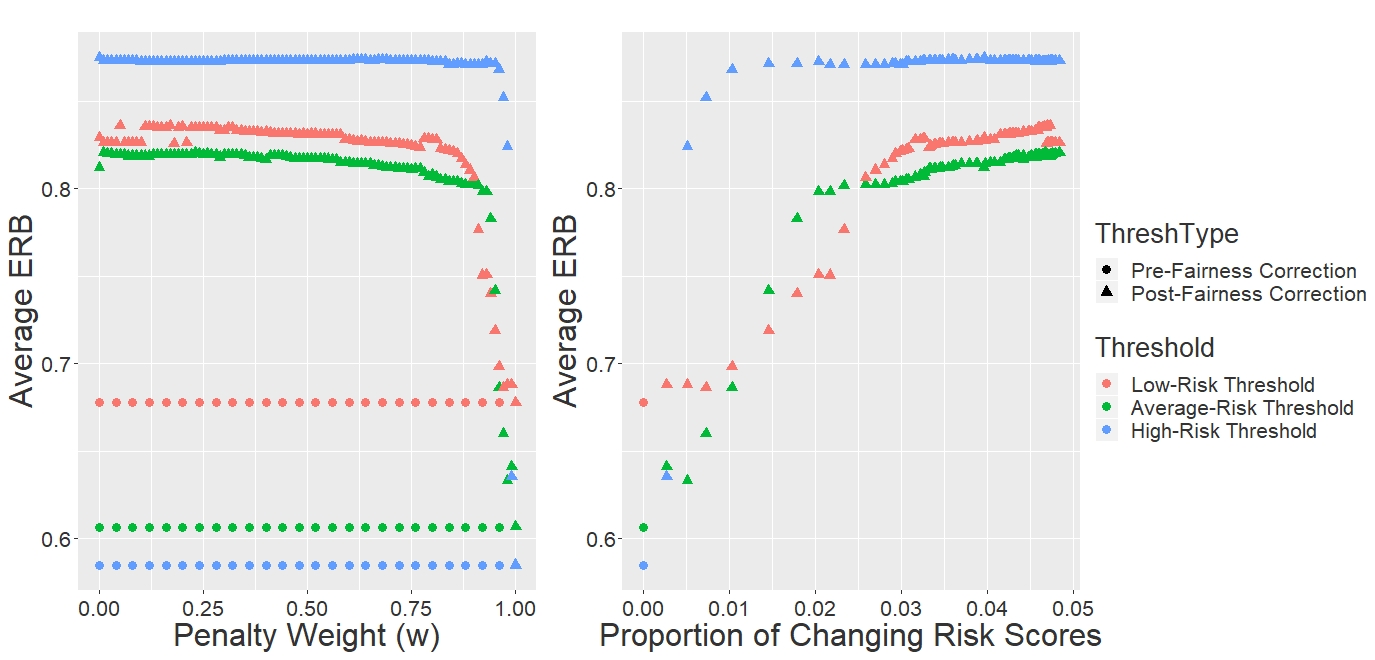}
    \caption{For each of the low-, average-, and high-risk thresholds, for both the pre- and post-fairness-corrected threshold values, the average Error Rate Balance versus the penalty weight, $w$, is plotted in the left-hand graphic, while the average Error Rate Balance versus the proportion of changing risk scores is plotted in the right-hand graphic.}
    \label{Figure:AvgERBvsPropChangingScores}
\end{figure*}

For all 101 $w$ values, the procedure described in Section \ref{SubSec:IdentifyingBestPenaltyWeight} was run with $J=200$.  For each of the three thresholds, for both the pre- and post-fairness-corrected threshold values, average Error Rate Balance versus the penalty weight, $w$, is plotted in the left-hand graphic of Figure \ref{Figure:AvgERBvsPropChangingScores}, while the average Error Rate Balance versus the proportion of changing risk scores is plotted in the right-hand graphic.  

Two observations stand out from Figure \ref{Figure:AvgERBvsPropChangingScores}.  First, fairness is not necessarily maximized when $w=0$, as evidenced for the low- and average-risk thresholds of the left-hand plot.  This is not unexpected since the optimization occurs within subsamples, as opposed to across subsamples, and facilitates treating $w$ as a tuning parameter.  Second, the relationship between Error Rate Balance and the proportion of changing risk scores, as evidenced in the right-hand plot, is different and non-linear across all three thresholds.  These trade-off continua reveal marginal gains in Error Rate Balance exist beyond 2.5\% of risk scores changing, with maximal gains requiring close to 5\% of risk scores changing.  Such a reality exemplifies the importance of understanding the functional relationship between fairness and accuracy. 
In fact, without such knowledge, it is unlikely that decision- and policy-makers will confidently identify within the available solution-space the "ideal" trade-off between accuracy and fairness.  Comparable plots of overall predictive performance measures versus penalty weight, are provided in Figure \ref{Figure:OvrlPPvsPenaltyWeight} of Appendix \ref{AppSub:OvrlPPvsPenaltyWeight}, and reveal analogous results; such findings support the use of the proportion of changing risk scores as a proxy for predictive performance at large.

\subsubsection{Resulting Post-Fairness-Corrected Threshold Values}
\label{SubSubSec:IdentifiedPostFCthresholdValues}
In light of the information conveyed through the plots of Figure \ref{Figure:AvgERBvsPropChangingScores}, and given that an upper bound of less than 5\% of changing risk scores is required to achieve maximum fairness, the "best" post-fairness-corrected threshold values were identified as those that maximize the extent to which Error Rate Balance is achieved.  Such maxima were identified 
at $w = 0.17,\; 0.23$, and $0.00$ for the low-risk, average-risk, and high-risk thresholds, respectively\footnote{For the low-risk threshold there are, in fact, multiple $w$-values where such a maximum is attained, and in theory this is also possible at other thresholds.  Hence, within the set of $w$-values yielding the maximum extent to which Error Rate Balance is achieved at a particular threshold, we select the one with the least variability in Error Rate Balance, as calculated in step 7 of Appendix \ref{App:IdentifyingBestPostFairnessCorrectedThresholdValuesStepByStep} and, if multiple $w$-values still qualify, we then select the largest $w$-value since it gives the greatest relative consideration to accuracy.}.  The corresponding set of pre- and post-fairness-corrected threshold values are given in Table \ref{Tbl:PreFCandPostFCthresholdValues}.
\begin{table*}[t]
    \caption{The pre-fairness-corrected (Pre-FC) and post-fairness-corrected (Post-FC) threshold values for all four protected attribute levels are provided below for each of the low-risk, average-risk, and high-risk thresholds.}
    \label{Tbl:PreFCandPostFCthresholdValues}
    \begin{tabular}{c|cc|cc|cc}
    \toprule
    \textbf{Protected Attribute} & \multicolumn{2}{c|}{\textbf{Low-Risk Threshold}} & \multicolumn{2}{c|}{\textbf{Average-Risk Threshold}} & \multicolumn{2}{c}{\textbf{High-Risk Threshold}} \\
    \textbf{Level}     & Pre-FC & Post-FC & Pre-FC & Post-FC & Pre-FC & Post-FC \\ \midrule
    BL        & 0.0420 & 0.0460 & 0.1275 & 0.1457 & 0.3494 & 0.3872 \\
    HPA       & 0.0420 & 0.0323 & 0.1275 & 0.1100 & 0.3494 & 0.3129 \\
    NV        & 0.0420 & 0.0484 & 0.1275 & 0.1352 & 0.3494 & 0.3757 \\
    WH        & 0.0420 & 0.0408 & 0.1275 & 0.1243 & 0.3494 & 0.3521 \\
    \bottomrule
    \end{tabular}
\end{table*}
From this table, it is evident at each threshold that more evidence\footnote{Which is to say a higher predicted probability.} is required of BL and NV child-reunification pairs than of HPA and WH reunification pairs before elevating the corresponding risk score.  Additional discussion related to these threshold values, along with a corresponding visualization, is provided in Appendix \ref{AppSub:VisualizingThresholds}.

\subsection{Error Rate Balance and Predictive Performance Assessment}
\label{SubSec:ERBandPPresults}
Given that a different $w$-value was utilized at each of the three thresholds, the procedure described in Section \ref{SubSec:IdentifyingBestPenaltyWeight} was run one additional time with 200 random subsamples (i.e., $J=200$) and the identified threshold-specific $w$-values.  Across these 200 subsamples, the average number of risk scores that changed in transitioning from the pre- to post-fairness-corrected threshold values was 4.76\%\footnote{Though we do not do so in this paper, this overall change percentage can be broken down by protected attribute level and by risk score.  Such information provides, for example, the fraction of BL child-reunification pairs assigned a 2 under the pre-fairness-corrected threshold values that were then assigned either a 1, 2, 3, or 4 under the post-fairness-corrected threshold values.}.  The average extent to which Error Rate Balance was achieved at each threshold, both for the pre- and post-fairness-corrected threshold values, is provided in Table \ref{Tbl:ERBresults}, along with the corresponding standard deviations.
\begin{table*}[t]
    \caption{The average extent to which Error Rate Balance is achieved at each threshold, both for the pre- and post-fairness-corrected threshold values, is provided, along with the corresponding standard deviations.}
    \label{Tbl:ERBresults}
    \begin{tabular}{c|cc|cc}
    \toprule
    & \multicolumn{2}{c|}{\textbf{Pre-Fairness-Corrected}} & \multicolumn{2}{c}{\textbf{Post-Fairness-Corrected}}\\
    Threshold & $\overline{\mbox{ERB}}(\Phi_{\cdot, w}, \vect{X})$ & $\text{SD}(\mbox{ERB}(\Phi_{\cdot, w}, \vect{X}))$ & $\overline{\mbox{ERB}}(\widehat{\vect{\theta}_{\cdot, w}}, \vect{X})$ & $\text{SD}(\mbox{ERB}(\widehat{\vect{\theta}_{\cdot, w}}, \vect{X}))$ \\
    \midrule
    Low-Risk  & 0.68 & 0.05 & 0.83 & 0.05 \\
    Average-Risk & 0.61 & 0.01 &0.82 & 0.03 \\
    High-Risk & 0.59 & 0.03& 0.87& 0.02 \\
    \bottomrule
    \end{tabular}
\end{table*}
From this table, it is clear that the fairness-correction procedure has meaningfully increased fairness at each threshold.  For example, at the high-risk threshold, the average Error Rate Balance increases from 0.59 to 0.87, which means that at this threshold, the most egregious disparity in either the false positive or false negative error rates between any two levels of the protected attribute has been reduced from 0.59 to 0.87.  The corresponding protected-attribute-level-specific average false negative and false positive rates are provided in Table \ref{Tbl:PAspecificFNRsFPRs} in Appendix \ref{AppSub:PAspecificFPRsAndFNRs}.  

\begin{figure}
    \centering
    \includegraphics[width=\linewidth, height=3.0in]{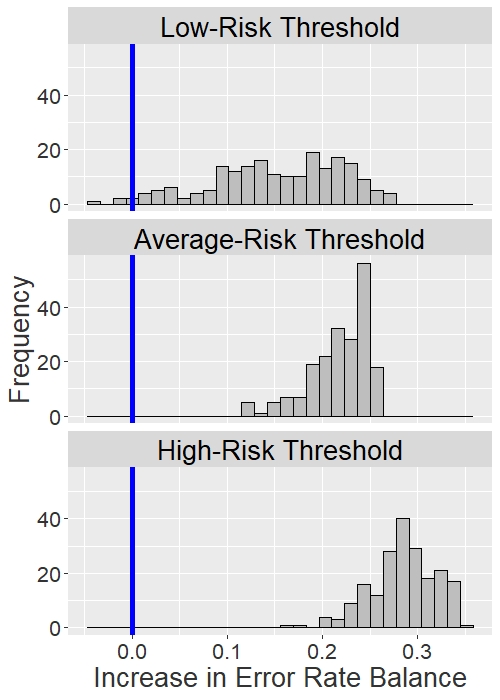}
    \caption{The distribution of the increase in the extent to which Error Rate Balance is achieved, at each threshold for 200 random subsamples, in transitioning from the pre- to post-fairness-corrected threshold values.  The vertical blue line in each plot represents the point at which no change occurred as a result of the procedure, whereas anything to the left of this line corresponds to decreased fairness and anything to the right corresponds to increased fairness.}
    \label{Figure:ERBchangeDistribution}
\end{figure}

To further demonstrate the improvements in Error Rate Balance resulting from the procedure, consider Figure \ref{Figure:ERBchangeDistribution}, where the distribution of this improvement, across the 200 subsamples, is plotted for each threshold. 
The vertical blue line in each plot of this figure corresponds to no change as a result of the procedure, while anything to the left of this line corresponds to decreased fairness and anything to the right corresponds to increased fairness.  From this figure, it is evident that for the average- and high-risk thresholds, all 200 subsamples result in a sizeable increase in fairness, whereas for the low-risk threshold, all but a few subsamples result in an improvement in fairness.

To assess the impact to predictive performance incurred through the use of the post-fairness-corrected threshold values, consider Table \ref{Tbl:AccuracyResults}, where the average overall predictive performance values at each threshold, for both the pre- and post-fairness-corrected threshold values, are provided\footnote{Note that the standard deviations of these values are not reported here since all such values are less than 0.02.}.  
\begin{table*}[t]
    \caption{The average overall predictive performance values at each threshold, for both the pre-fairness-corrected (Pre-FC) and post-fairness-corrected (Post-FC) threshold values, are provided. These measures include Accuracy (ACC), False Negative Rate (FNR), False Positive Rate (FPR), Negative Predictive Value (NPV), and Positive Predictive Value (PPV).}
    \label{Tbl:AccuracyResults}
    \begin{tabular}{c|cc|cc|cc}
    \toprule
    Performance & \multicolumn{2}{c|}{\textbf{Low-Risk Threshold}} &  \multicolumn{2}{c|}{\textbf{Average-Risk Threshold}} &  \multicolumn{2}{c}{\textbf{High-Risk Threshold}}  \\
    Measure & Pre-FC & Post-FC & Pre-FC & Post-FC & Pre-FC & Post-FC  \\ \midrule
    ACC & 0.46 & 0.45 & 0.71 & 0.70 & 0.83 & 0.83  \\ 
    FNR & 0.13 & 0.13 & 0.38 & 0.37 & 0.75 & 0.75  \\ 
    FPR & 0.62 & 0.64 & 0.27 & 0.28 & 0.05 & 0.05  \\ 
    NPV & 0.93 & 0.93 & 0.90 & 0.91 & 0.86 & 0.86  \\ 
    PPV & 0.22 & 0.22 & 0.32 & 0.32 & 0.52 & 0.52  \\
    \bottomrule
    \end{tabular}
\end{table*}
From this table, it is readily apparent that the accuracy costs associated with the substantive gains in algorithmic fairness are comparatively marginal.  In fact, when rounded up to two decimal places, the overwhelming majority of such measures are unchanged as a result of the procedure.  The biggest cost for any measure at any of the three thresholds is the roughly two percentage point increase in the false positive rate at the low-risk threshold (i.e., 0.62 to 0.64), which we view as an acceptable cost relative to the substantive gains achieved in algorithmic fairness across all three thresholds.

\subsubsection{Considering Alternative Group-Level Definitions}
\label{SubSubSec:ResultsAltDefns}
\begin{figure}
    \centering
    \includegraphics[width=\linewidth, height=3.0in]{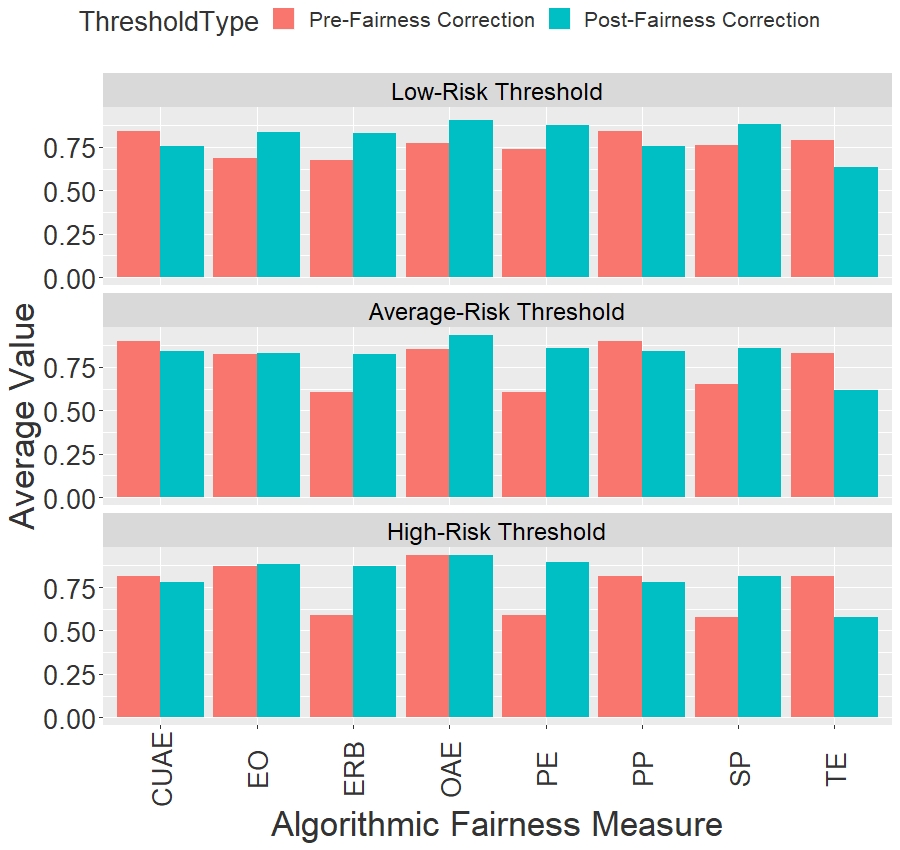}
    \caption{At each of the low-, average-, and high-risk thresholds, for both the pre- and post-fairness-corrected threshold values, the corresponding average value of eight group-level definitions of fairness are plotted.  These definitions include Conditional Use Accuracy Equality (CUAE), Equality of Opportunity (EO), Error Rate Balance (ERB), Overall Accuracy Equality (OAE), Predictive Equality (PE), Predictive Parity (PP), Statistical Parity (SP), and Treatment Equality (TE).}
    \label{Figure:AllFairnessDefsPlots}
\end{figure}

\begin{figure}
    \centering
    \includegraphics[width=\linewidth, height=1.5in]{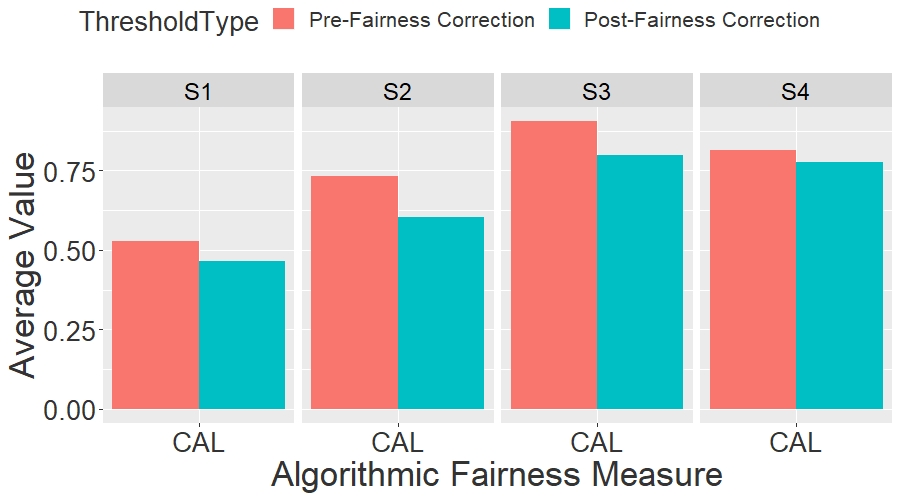}
    \caption{At each of the four possible scores (i.e., S1, S2, S3, and S4), for both the pre- and post-fairness-corrected threshold values, the corresponding average value of the Calibration (CAL) group-level definition of fairness is plotted.}
    \label{Figure:CALfairnessDefsPlots}
\end{figure}

To audit the algorithmic fairness of the pre- and post-fairness corrected threshold values provided in Table \ref{Tbl:PreFCandPostFCthresholdValues} under eight alternative group-level definitions, the process described in Section \ref{SubSec:ERBandPPresults} and thoroughly detailed in Appendix \ref{App:IdentifyingBestPostFairnessCorrectedThresholdValuesStepByStep} was utilized\footnote{Technically, these measures were calculated at the same time the corresponding Error Rate Balance measures and predictive performance measures were calculated.}.  In Figure \ref{Figure:AllFairnessDefsPlots}, the average value of each of eight group-level definitions of algorithmic fairness is plotted across each threshold, for both the pre- and post-fairness-corrected threshold values.  
Similarly, in Figure \ref{Figure:CALfairnessDefsPlots}, the average value of the one remaining, subtly different, group-level definition (i.e., Calibration) is plotted for each risk score (i.e., denoted in increasing order of risk as S1, S2, S3, and S4), for both the pre- and post-fairness-corrected threshold values.  

From these plots, two observations stand out.  First, the fairness measures for Calibration, Predictive Parity, and Conditional Use Accuracy Equality have all decreased.  This is an unavoidable consequence of the results referred to as Impossibility Theorems \cite{BHJKR2017}.  In particular, these results dictate that for all realistic\footnote{More specifically, as long as the prevalence of the outcome differs across the levels of the protected attribute and as long as a perfect classifier does not exist \cite[][pg. 19]{BHJKR2017}, the corresponding proofs hold.  In the context of this paper, such differing outcome prevalence is demonstrated in Figure \ref{Figure:PAdistPrevByPAlevel} of Appendix \ref{App:IdentfyingPA}.} applications, Error Rate Balance cannot simultaneously be achieved with either Conditional Use Accuracy Equality\footnote{Technically, the proofs pertain to Predictive Parity, but as it represents "half" of Conditional Use Accuracy Equality, the results still hold.} or Calibration \cite{Chouldechova2017, KMR2016}.  Interestingly, Treatment Equality has also markedly decreased at all three thresholds.  Second, while the biggest improvements in fairness as a result of applying the procedure are unsurprisingly seen with Error Rate Balance\footnote{Increases in Equal Opportunity and Predictive Equality are also seen, which is not surprising given that these two definitions collectively make up Error Rate Balance.}, both the Statistical Parity and Overall Accuracy Equality measures, though not directly corrected for in the procedure, have also increased at each threshold.  Whether such positive associations are indicative of an underlying theoretical relationship, or an artifact of this use-case, is beyond the scope of this manuscript. 

Undoubtedly, some of these observed trade-offs in fairness are difficult to accept, but it is important to remember that they are an unavoidable consequence of increasing the extent to which Error Rate Balance is achieved.  For jurisdictions that value an alternative definition of algorithmic fairness, the procedure can still effectively be applied.  In Appendix \ref{App:DemonstratingGeneralizability}, we demonstrate such generalizability by successfully applying the procedure under three alternative specifications of algorithmic fairness: Conditional Use Accuracy Equality, Treatment Equality, and Calibration. 

Regardless of the identified definition, however, such difficult trade-offs will persist.  And, in fact, as demonstrated in Figures \ref{Figure:AllFairnessDefsPlots} and \ref{Figure:CALfairnessDefsPlots}, these trade-offs are being made even when no fairness-correction procedure is utilized.  Importantly, these difficult trade-offs exist even without the use of an algorithm, and "[r]ejecting model-driven or automated decision making is not a way to avoid these problems" \cite[][pg. 15]{MPBDL2020}.  Hence, rather than accepting the "default" trade-offs of either the status quo or the uncorrected classification algorithm, we advocate for identifying the trade-offs that are most appropriate for the use-case, prior to training the algorithm, and then fairness-correcting accordingly.

\section{Discussion}
\label{Sec:FinalThoughts}
We conclude this manuscript with a discussion of the business-case utility of the fairness-corrected risk scores, along with some of the limitations of the presented correction procedure.
\subsection{Utility of the Fairness-Corrected Reunification Algorithm for Decision Support} 
\label{SubSec:AssessingUtilityOfRiskScores}
While the AUC of the algorithm, as well as the standard assortment of predictive performance measures for the corresponding fairness-corrected risk scores, suggest a viable predictive analytic tool, the true value of the tool is best recognized through alternative use-case specific metrics (Table \ref{Tbl:CalibrationTable}).  Ultimately, such a tool can help maximize the amount and success rate of reunifications, as well as identify high risk reunifications which may benefit from supportive services and resources.  Importantly, in implementing this risk algorithm to inform reunification decisions, the understandably and unavoidably inconsistent and incomplete use of administrative data by permanency workers is buttressed with a decision support tool that not only leverages administrative data in a consistent and thorough manner, but also adheres to an agreed upon set of shared values articulated through the identified definition of fairness.

\subsection{Setting a New Standard for Algorithmic Fairness within Child Welfare}
While the described correction procedure represents an improvement upon the current "standards" of algorithmic fairness in the predictive analytic tools of Child Welfare, it is far from perfect.  We highlight here one such imperfection.  In particular, because we have not explored every proposed procedure seeking to mitigate, or ``correct'' for, a lack of algorithmic fairness, it is likely that the utilized procedure does not represent in general, or in this use-case, the optimal approach.  Pre- and in-processing approaches that are not appropriate for this use-case may be appropriate for, or more optimal with, other use-cases, or for other Child Welfare jurisdictions embracing different values\footnote{Such as, for example, a different definition of algorithmic fairness.} within this use-case.   Because it is possible to combine approaches across pre-, in-, and post-processing classes \cite{BHJKR2017}, it may even be the case that the true optimal solution comes in the form of a composite correction procedure.  The pursuit of a more optimal solution, however, represents an ideal future-state for the predictive algorithms of Child Welfare; our objective for the time being is to establish a baseline for such a pursuit.

Despite this and other unmentioned imperfections, the presented procedure is unquestionably effective at increasing algorithmic fairness, according to a variety of group-level definitions, with comparatively minimal cost to accuracy.  We therefore anticipate that such a procedure could be useful and valuable in any number of other contexts outside of Child Welfare.  In such use-cases, the observational units may be independent and the utilized subsampling scheme, as well as the corresponding computational burden, largely unnecessary\footnote{This subsampling scheme can straightforwardly be circumvented by setting $I=1$ in the procedure detailed in Appendix \ref{App:TheFairnessCorrectionProcedureStepByStep}}.  We suspect, however, that in such instances it may instead be beneficial to deploy a bootstrap resampling scheme\footnote{Such a scheme can seamlessly replace the subsampling scheme in the code.}, which should yield more robust post-fairness-corrected threshold values relative to a single iteration of the optimizer across the entire sample.  Regardless, we hope that this work both broadly furthers the efforts surrounding algorithmic fairness in Child Welfare and meaningfully provides an additional fairness-correction procedure for algorithm developers and stakeholders.



\bibliographystyle{ACM-Reference-Format}
\bibliography{main}

\appendix

\section{Two Misguided Conceptions Of Algorithmic Fairness}
\label{App:MisguidedConception}
In doing this work and in broadly sharing it, we have learned to call attention to two common, but misguided conceptions of what constitutes algorithmic fairness.  
\begin{enumerate}
    \item Removing the protected attribute from the feature set during algorithm training.
    \item Verifying the protected attribute is not statistically significant in the model.
\end{enumerate} Both of these conceptions are akin to a "fairness through unawareness" approach, but both ultimately fail to acknowledge the likely relationships that exist between the protected attribute and the other features accessible to the algorithm.  Berk et. al. phrase it this way: ``Even when direct indicators of protected group membership, such as race and gender, are not included as predictors, associations between these measures and legitimate predictors can ``bake in'' unfairness'' \cite[][pg. 2]{BHJKR2017}.  Hardt et. al., referencing \cite{PRT2008}, convey it this way: "...this idea of "fairness through unawareness" is ineffective due to the existence of \textit{redundant encodings}, [or] ways of predicting protected attributes from other features" \cite[][pg. 1]{HPS2016}.  In particular, the first conception problematically has not removed from the feature set any proxies for the protected attribute, while the second has problematically only verified that the protected attribute is not significant \textit{after accounting for the effects of all other variables in the model}.  Hence, algorithmic fairness is not achieved by either (only) removing the protected attribute from the feature set or by only ``verifying'' its lack of statistical significance.

\section{Modeling Best Practices}
\label{App:ModelBestPractices}
\subsection{Selective Labeling}
\label{AppSubSec:SelectiveLabeling}
The selective labeling problem \cite{lakkaraju2017selective} occurs when historical data are overly influenced by the very decision which a tool seeks to support. In Child Welfare reunification, this occurs because outcomes (e.g., whether the child experiences further abuse/neglect at home) is partly influenced by the very act of returning the child home. Thus, the historical data should be analyzed solely for child-transition pairs which represent a reunification with family. Thus, the outcomes of interest become conditional on the decision, and the machine learning classifier is less likely to perpetuate poor decisions. In this way, when we refer to a child welfare outcome, we are sure to avoid bias via the selective label problem by ensuring that we are calculating the likelihood of the outcome conditional on the decision (e.g., the likelihood of a return to substitute care conditional on the child’s reunification with his/her family).
\subsection{Repeated Observations}
\label{AppSubSec:RepeatedObs}
Repeated observation leakage can occur if independence is violated in the data used to train the machine learning classifier. In Child Welfare reunification, children may have been involved in multiple reunification events over time. Thus, it is important to unduplicate the historical data so that the same child does not occur twice in the data set. When the data is split into training data and testing data, in order to test the predictive performance of the machine learning classifier, this unduplication ensures that information does not “leak” between the training set and the testing set, which can cause predictive performance to be inflated. Because the unduplication reduces the volume of the historical data, we have constructed a repeated sample technique which repeats the unduplication step over and over until a series of classification models are constructed and all child-reunification pairs have been included in at least one of the models. This technique is rooted in the common data science practice of model ensembling. It ensures that we can maximally draw on historical data in the construction of our decision support tool, and do so in a way that prevents the inflation of predictive performance metrics.
\subsection{Censoring and Data Windows}
\label{AppSubSec:CensoringWindows}
Oregon Child Welfare’s SACWIS system came online in August 2011. This means that historical data is available in a reliable and consistent format only from a certain historical time point. Reunifications that occurred in February 2012 have only six months of reliable historical data, while reunifications that occurred in August 2019 have eight years of reliable historical data. For this reason, it is important to standardize the historical data window. Otherwise, the date of a historical reunification will have undue influence on the historical calculation of risk, and the model will not be generalizable to new observations. To prevent this from occurring, we set a historical data window to 1.5 years.

A similar censoring bias can occur with the outcome window. For example, a child involved in a February 2012 reunification has eight years of time in which to experience (or not experience) an outcome, whereas a child involved in an August 2019 reunification has had only one year to experience (or not experience) the outcome. Consequently, a fixed outcome window of one year was established to prevent undue data bias via the timing of the report.
\section{Identifying The Protected Attribute}
\label{App:IdentfyingPA}
Within the utilized administrative data, the potential child features available for use as a protected attribute include race and ethnicity\footnote{With 6 possible levels: Asian, Black, Hispanic, Native American or Alaskan Native, Pacific Islander, and White}, sex \footnote{With 2 possible levels male or female}, and ICWA-status\footnote{The Indian Child Welfare Act status of a child is best understood as a binary federal designation, as opposed to a race- or ethnicity-based identifier}.  Given that no evidence of a lack of algorithmic fairness across sex was evident either pre- or post-fairness-correction procedure, the identification of the protected attribute focused on race and ethnicity, as well as ICWA-status.  Because sample sizes across some of these groups/designations are too small to enable use of the procedure described in Section \ref{Sec:FairnessCorrectionProcedure} and detailed in Appendix \ref{App:TheFairnessCorrectionProcedureStepByStep}, combining groups was necessary.  Each of these small-in-sample-size groups are merged with a group that is most similar in outcome prevalence.  This approach yielded a protected attribute with four levels: Black (BL); Hispanic, Pacific Islander, or Asian (HPA); Native American or ICWA-eligible (NV); and White (WH).

\begin{figure}
    \centering
    \includegraphics[width=\linewidth, height=3.0in]{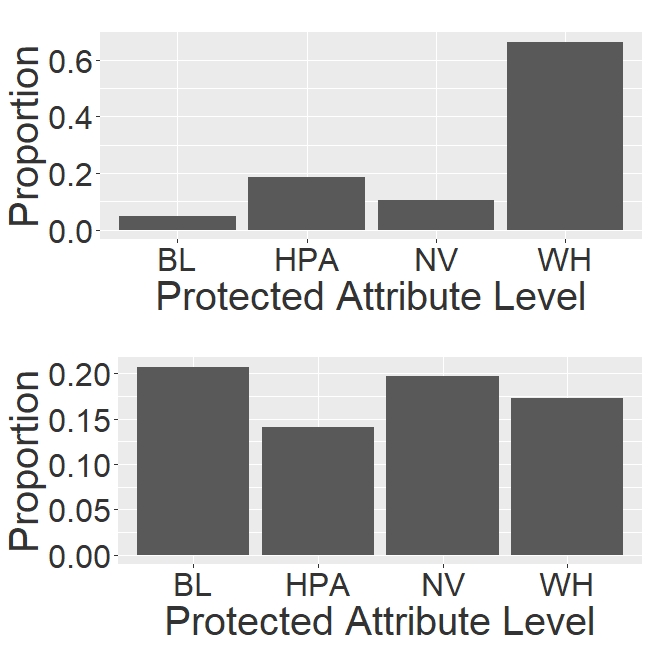}
    \caption{The average distribution of the race and ethnicity-based protected attribute across 200 random subsamples is plotted in the top graphic, while the corresponding average prevalence of the outcome, by protected attribute level, is plotted in the bottom graphic.}
\label{Figure:PAdistPrevByPAlevel}
\end{figure}

In Figure \ref{Figure:PAdistPrevByPAlevel}, the average distribution of this four-level protected attribute, across 200 random subsamples\footnote{Since a child can have multiple reunifications, each subsample randomly selects a single reunification event for each child in the data set.}, is provided in the top graphic, while the corresponding average prevalence of the outcome across the levels of this protected attribute is provided in the bottom graphic.  From the top plot, it is apparent that children identifying as WH make up the overwhelming majority of all child-reunification pairs at around 66\%, on average, while children identifying as BL, HPA, and NV make up, on average, approximately 5\%, 19\%, and 10\% of the overall child-reunification pairs, respectively.  From the bottom plot, on the other hand, it is apparent that children identifying as BL, as well as children identifying as NV, are disproportionately likely to experience the adverse event.  More specifically, with an average prevalence of approximately 20.74\% and 19.66\%, respectively, BL and NV child-reunification pairs are approximately 47\% and 39\% more likely, on average, to experience the adverse event than HPA child-reunification pairs, among whom the average prevalence is approximately 14.10\%.  Similarly, though less amplified, BL and NV child-reunification pairs are approximately 20\% and 14\% more likely, on average, to experience the adverse event than WH child-reunification pairs, among whom the average prevalence is approximately 17.25\%.

\section{Why Use a Group-Level Definition?}
\label{App:WhyGroupLevel}
Following the structure and content provided in \cite{VR2018}, we discuss below the three main categories of definitions of algorithmic fairness and why we ultimately focused on the category of group-level definitions.

Most definitions of algorithmic fairness seem to fall into three identified categories: group-level, individual-level, and causal reasoning-based.  Within group-level definitions, observational units\footnote{Child-reunification pairs for the reunification algorithm.} are grouped according to one or more protected attributes (e.g., race, ethnicity, gender, etc.) and some measure, typically of predictive performance, is calculated and compared across these groupings.  If the resulting measure is equal across all groups, then the algorithm is deemed fair; otherwise, the algorithm is deemed not fair. Criticisms surrounding this category of definitions are 1) any unfairness across non-protected attributes\footnote{As well as other protected attributes not directly incorporated into the groupings.} could be missed and 2) observations which are ``close'' to one another in feature space can still be unfairly ``far'' from one another in prediction space.

These criticisms give rise to the next category of definitions, individual-level, where the guiding principle is that observations which are ``close'' to one another in feature space should be ``close'' to one another in prediction space.  What is meant by ``close'' depends on the particular definition.  For example, under Causal Discrimination,  ``close'' means ``identical''; i.e., observational units with identical feature vectors should have identical predictions \cite{VR2018}.  Another example, Fairness Through Awareness, is a relaxation on ``identical'', where ``close'' now means ``no further than''; i.e., observational units can be no further apart in prediction space than they are in feature space \cite{DHPRZ2011}. 

While such a ``closeness'' principle resonates for us at a conceptual level, we have found it difficult to embrace or utilize at a practical level.  In particular, two feature vectors can be identified as ``close'' as a result of biases, rather than any ``true'' underlying similarity.  For example, two child-reunification pairs may have very similar allegation histories, but such similarities may be attributable to reporting biases among the public, rather than any true commonality between those children and their respective families.  Hence, any use of this category of definitions within an algorithmic fairness procedure should, we believe, necessitate the use of a pre-processing component, which lacks certain functionality specifications we value for this and other use cases, as discussed in Appendix \ref{App:IdentifyingFairnessCorrectionProcedure}.  Additionally, the identification of an appropriate distance metric appears at this time to be a challenging and subjective-laden endeavor.  Perhaps such observations contributed to Berk et. al. stating that individual-level definitions of fairness  ``...currently cannot be operationalized in a useful manner'' \cite[][pg. 15]{BHJKR2017}. In the end, these concerns collectively lead us to look to alternative categories of definitions.

The third and final category of definitions is causal reasoning-based.  While there are numerous specifications within this category, including Counterfactual Fairness, No Unresolved Discrimination, No Proxy Discrimination, and Fair Inference, we have neither explored nor seriously considered such definitions.  The justification for this sweeping dismissal resides solely in the causal graph that is foundational to these definitions.  In particular, we are in no way comfortable with or confident in our ability to create a graph depicting the causal relationships between the outcome and the more than 500 features utilized in the algorithm.  Furthermore, even if we were comfortable with and confident in our ability to construct such a graph, it would be ``...impossible to test an existing classifier against causal definitions of fairness'' \cite[][pg. 6]{VR2018}.  Hence, while such an approach to defining algorithmic fairness may represent an ideal, currently it is not a pragmatically viable option.

By process of elimination, the category of group-level definitions remains.  While this category is clearly imperfect, the options within it are valuable, appropriate, and useful.  Additionally, this category of definitions allows for the aforementioned functionality specifications, discussed in Appendix \ref{App:IdentifyingFairnessCorrectionProcedure}, which we value for this and other use cases.  Ultimately, ``settling'' for this category of algorithmic fairness can be viewed as one example of the imperfectness of the currently and pragmatically available options, but it should also be recognized that such ``settling'' is unquestionably an improvement upon the current standard.

\section{Why Error Rate Balance?}
\label{App:WhyErrorRateBalance}
There are many proposed group-level definitions of algorithmic fairness.  Below we discuss the vast majority of such options covered in \cite{VR2018} and provide the logic supporting the decision to utilize Error Rate Balance\footnote{Note that when met, Error Rate Balance is equivalent to Equalized Odds.}.  While we, along with the algorithm's stakeholders, have identified this definition as the ``best'' option, other jurisdictions developing such a tool may appropriately identify as "best" an alternative group-level definition.  

\subsection{Easier to Dismiss Definitions}
\label{AppSubSec:EasierToDismissDefns}
A commonly proposed group-level definition is Statistical Parity.  This definition asserts that at a given threshold, the proportion of children predicted\footnote{More precisely, the proportion of child-reunification pairs with a predicted probability above the threshold.} to return to substitute care within one year\footnote{Note that such a dichotomous assessment holds when only a single threshold is utilized.  When multiple thresholds are utilized, as is the case here, then each subsequent threshold is viewed as an escalation of risk, rather than an on/off switch.  Regardless, it is helpful thinking about each threshold in this manner so as to facilitate understanding around how the measure is calculated.} should be equal across all levels of the protected attribute.  In other words, the predicted outcome prevalence should be uniformly distributed.  While criticisms of Statistical Parity are extensively discussed in \cite{DHPRZ2011}, our central concern for this use-case is grounded in the likely reality that ``...it can lead to undesirable decisions...'' \cite[][pg. 13]{BHJKR2017}.  In particular, translating the criticism articulated in \cite{HPS2016} into the context of the reunification algorithm, imagine that for one level of the protected attribute reunifications are only attempted with "qualified" child-reunification pairs, but for one or more other levels of the protected attribute reunifications are attempted with both "qualified" and "unqualified" child reunification pairs.  In such a situation, as long as the proportion of attempted reunifications match across the levels of the protected attribute, statistical parity would not raise an alarm.  Such a lack of discretion, when coupled with the undeniable existence and corresponding ramifications of sample bias, renders Statistical Parity, with its strict a priori demand for a uniform distribution, inappropriate for this use-case.

A second definition is Overall Accuracy Equality, which at a given threshold asserts that the accuracy of the algorithm should be equal across the levels of the protected attribute.  In the context of this use-case, this means that the proportion of times the algorithm either correctly labels a child-reunification pair as successful\footnote{In other words, the predicted probability is lower than the threshold and the child does not return to substitute care within 1 year.}, or correctly labels a child-reunification pair as unsuccessful\footnote{In other words, the predicted probability is greater than the threshold and the child does return to substitute care within 1 year.}, should be the same across all four levels of the protected attribute.  Our avoidance of this definition lies in the criticism offered in Berk et. al.: ``Overall accuracy equality is not commonly used because it does not distinguish between accuracy for successes and accuracy for failures''  \cite[][pg. 13]{BHJKR2017}.  In the context of the reunification algorithm, this means that Overall Accuracy Equality would not raise an alarm if the algorithm was better at predicting ``failed'' reunifications for some levels of the protected attribute and better at predicting ``successful'' reunification for other levels of the protected attribute, so long as the overall correctness ``evened out'' across these two scenarios.  Such an inability was not something we or the algorithm's stakeholders were willing to accept in light of other available options. 

\subsection{Valued but ``Incomplete'' Definitions}
\label{AppSubSec:ValuedButIncompleteDefns}
The next three definitions -- Predictive Parity, Predictive Equality, and Equal Opportunity -- were all dismissed for the same reason.  In particular, all three only address ``half'' of the situation.  These definitions may ultimately be appropriate in contexts where it is reasonable to assume each and every observational unit desires the same outcome, such as in academic admissions (i.e. admittance into the school) or loan applications (i.e. qualification for the loan), but such knowledge cannot safely be assumed when the observational units are child-reunification pairs. More specifically, whether the most appropriate setting for the child is via reunification or via substitute care varies across children.  With this in mind, we highlight the ``incompleteness'' of each definition in turn.  

Predictive parity asserts that at a given threshold the positive predictive value (i.e., PPV) of the algorithm must be equal across all levels of the protected attribute.  This means that when the algorithm predicts a child-reunification pair will ``fail'', the likelihood that it actually does ``fail'' is equivalent across the levels of the protected attribute.  Such a standard, however, requires nothing of the negative predictive value (i.e., NPV), which is the counterpart to PPV.  This means that, under Predictive Parity, when the algorithm predicts a child-reunification pair will ``succeed'', the likelihood that it actually does ``succeed'' can depend on the level of the protected attribute.   Hence, choosing such a definition, in this context, leaves much to be desired.  Given that an alternative, more ``complete'', definition is available, as discussed in Appendix \ref{AppSubSec:DifficultToDismissDefinitions}, there is also no compelling reason to choose Predictive Parity. 

The remaining two definitions serve as counterparts to one another.  In particular, Predictive Equality asserts that at a given threshold the false positive rate (i.e., FPR) must be the same for all levels of the protected attribute, while Equal Opportunity asserts that the false negative rate (i.e. FNR) must be the same.  In other words, Predictive Equality requires that the probability of a ``successful'' child-reunification pair being incorrectly predicted to ``fail'' be the same for each level of the protected attribute, while Equal Opportunity requires that the probability of a ``failed'' child-reunification pair being incorrectly predicted to ``succeed'' be the same.  Having to choose one of these definitions over the other is thus clearly dissatisfying.  However, given that an alternative definition is available which encompasses both, as discussed in Appendix \ref{AppSubSec:DifficultToDismissDefinitions}, there is no need to make such a choice.  

\subsection{Error Rate Balance and Alternative, Difficult To Dismiss, Definitions}
\label{AppSubSec:DifficultToDismissDefinitions}
In addition to Error Rate Balance, three definitions to address remain: Conditional Use Accuracy Equality, Treatment Equality, and Calibration.  Conditional Use Accuracy Equality was alluded to above in Appendix \ref{AppSubSec:ValuedButIncompleteDefns}, and is simply the extension of Predictive Parity to additionally require that the NPV matches across the levels of the protected attribute.  Hence, when the algorithm makes a prediction for a child-reunification pair, be it either of "failure" or of "success", Conditional Use Accuracy Equality requires that the likelihood of actually observing the outcome be the same across the levels of the protected attribute.   

Treatment Equality, on the other hand, asserts that at a given threshold the ratio of false positives to false negatives must be the same across the levels of the protected attribute.  In other words, this definition requires that the number of ``successful'' child-reunification pairs incorrectly predicted to ``fail'', relative to the number of ``failed'' child-reunification pairs incorrectly predicted to ``succeed'', be the same for all levels of the protected attribute.  Observe that such a ratio at a specified threshold, for each protected attribute level, is ultimately measuring the relative importance of a false positive to a false negative mistake.  For example, if such a ratio has a value of two at a specified threshold, then false negatives are treated as the more "costly" mistake since each false negative comes at the expense of two false positives.  Hence, regardless of the value of this ratio at a threshold, Treatment Equality requires that this relative cost value be the same across all levels of the protected attribute.

The last alternative definition, Calibration\footnote{Other names given to this definition include Test-Fairness and Matching Conditional Frequencies \cite{VR2018}.}, is subtly, but importantly, different from the other group-based definitions discussed in this manuscript.  In particular, it asserts that for a given risk score\footnote{Rather than at a given threshold.  Note that a risk score can essentially be viewed as a "bin" of predicted probabilities}, the proportion of observational units that ultimately experience the adverse event must be the same across the levels of the protected attribute.  Stated differently, Calibration requires the amount of risk associated with a given score to be the same across the levels of the protected attribute.  In the context of the reunification algorithm, this means that when the algorithm assigns a child-reunification pair a risk score of $S$\footnote{Where $S\in\{S1,S2,S3,S4\}$}, the likelihood that the reunification actually does ``fail'' is equivalent across the four protected attribute levels\footnote{Note that this definition is related to, but distinct from, Well-Calibrated, which is yet another group-level definition of algorithmic fairness that is not formally addressed in this manuscript.  We did not formally consider Well-Calibrated, which requires the algorithm to be calibrated in both the fairness sense just defined, as well as in the traditional sense of a calibrated statistical model (i.e., the predicted probability of experiencing the adverse event should be equal to the observed proportion of adverse events among observations assigned to that risk score), because our purposes with the predictive algorithm are sufficiently served regardless of whether the probabilities are in fact statistically calibrated.}.  It is also worth noting here, that while Calibration is different from Predictive Parity, risk scores that satisfy Calibration will also necessarily satisfy Predictive Parity, though only in the event of a two-level risk scoring system \cite{Chouldechova2017}.

Finally, we address Error Rate Balance, which is equivalent to Equalized Odds, as defined in \cite{HPS2016}.  This definition of fairness is ultimately a coupling of two of the "incomplete" definitions discussed in Appendix \ref{AppSubSec:ValuedButIncompleteDefns}: Predictive Equality and Equal Opportunity.  More specifically, at a particular threshold, Error Rate Balance requires that the false positive rate (i.e., FPR) be the same across the levels of the protected attribute, as well as the false negative rate.  Hence, given an outcome for a child-reunification pair, be it "success" or "failure", Error Rate Balance requires that the probability of a corresponding incorrect prediction label be the same across the levels of the protected attribute.  Stated differently, this definition recognizes that the algorithm, like all algorithms, is going to make mistakes, but it ultimately requires that those mistakes be proportionately experienced across the levels of the protected attribute.

\subsection{The Value of Error Rate Balance}
\label{AppSubSec:ValueErrorRateBalance}
There are undoubtedly compelling arguments for utilizing any of the four definitions of Appendix \ref{AppSubSec:DifficultToDismissDefinitions}, which begs an important question: Could another definition of algorithmic fairness be created which encompasses all four?  After all, Error Rate Balance and Conditional Use Accuracy Equality are both examples of such composite definitions.  Unfortunately, the answer to this question is a definitive no.  In particular, it has been proven that for all realistic\footnote{More specifically, as long as the prevalence of the outcome differs across the levels of the protected attribute and as long as a perfect classifier does not exist \cite[][pg. 19]{BHJKR2017}, the corresponding proofs hold.  In the context of this paper, such differing outcome prevalence is demonstrated in Figure \ref{Figure:PAdistPrevByPAlevel}.} applications, Error Rate Balance cannot simultaneously be achieved with either Conditional Use Accuracy Equality\footnote{Technically, the proofs pertain to Predictive Parity, but as it represents "half" of Conditional Use Accuracy Equality, the results still hold.} or Calibration \cite{Chouldechova2017, KMR2016}.  These results are referred to as Impossibility Theorems \cite{BHJKR2017} and it is likely the case that many more such impossibility theorems exist, encompassing other group-level definitions of algorithmic fairness \cite{N2018}.  The practical implications of this reality are non-trivial: ``...altering a risk algorithm to improve matters can lead to difficult stakeholder choices.  If it is essential to have conditional use accuracy equality, the algorithm will produce different false positive and false negative rates across the protected [attribute levels].  Conversely, if it is essential to have the same rates of false positives and false negatives across protected [attribute levels], the algorithm cannot produce conditional use accuracy equality.  Stakeholders will have to settle for an increase in one for a decrease in the other'' \cite[][pg. 19]{BHJKR2017}.  This is an unavoidably difficult trade-off that exist in addressing algorithmic fairness at the group level.

So why did we ultimately ``settle on'' Error Rate Balance?  Such a decision is rooted in the values that Error Rate Balance supports relative to Conditional Use Accuracy Equality, and by some measure of relatedness, Calibration\footnote{Technically, the fundamental criticism we raise here is with respect to Conditional Use Accuracy Equality, but it can also apply to Calibration depending on the distribution of predicted outcome prevalence across the levels of the protected attribute.}.  In particular, Error Rate Balance enforces that the prediction and the protected attribute are independent when conditioned on the outcome, which ``...encourages the use of features that allow to directly predict [the outcome], but prohibits abusing [the protected attribute] as a proxy for [the outcome]'' \cite[][pg. 3]{HPS2016}.  In contrast, Conditional Use Accuracy Equality enforces that the outcome and the protected attribute are independent when conditioned on the prediction, which equates to ``...utilizing the protected attribute for optimal predictive power, rather than protecting [against] discrimination based on it'' \cite[][pg. 15]{HPS2016}.  When viewed in this light, Error Rate Balance demonstrates greater alignment, relative to Conditional Use Accuracy Equality and Calibration, with the values we and our stakeholder's desire from a definition of algorithmic fairness.  It is important to recognize, however, that with a lack of Calibration, and by relatedness Conditional Use Accuracy Equality, the proportion of child-reunification pairs within each risk score that actually end up experiencing the adverse event will differ across protected attribute levels, which can "...have the unintended and highly undesirable consequence of incentivizing [tool users] to take [the protected attribute] into account when interpreting predictions" \cite[][pg. 1]{PRWKW2017}.  Ultimately, such a trade-off must be effectively addressed and managed in training users of the tool.

The lone remaining alternative to Error Rate Balance is then Treatment Equality.  Our decision to dismiss Treatment Equality is not rooted in a value-based argument, but instead in our uncertainty around 1) what it ultimately requires of the relationship between the outcome, the prediction, and the protected attribute and 2) why it would be preferable to Error Rate Balance.  That being said, if we encounter or are presented with information that convincingly supports the utilization of Treatment Equality over Error Rate Balance, we would act accordingly.  Until then, however, Error Rate Balance ``best'' represents what we are seeking in a definition of algorithmic fairness for this use-case.

\section{Quantifying Error Rate Balance}
\label{App:QuantifyingErrorRateBalance}
We describe and demonstrate below a measure for quantifying the extent to which Error Rate Balance is achieved\footnote{As all group-level definitions are technically met or unmet by the algorithm, we require a method for quantifying the extent to which the definition is achieved.  The measure presented in this manuscript is applicable with any of the other group-level definitions of algorithmic fairness discussed in Appendix \ref{App:WhyErrorRateBalance}}.  The measure we utilize for this purpose is rooted in the error rate ratios that are available across all possible pairings of protected attribute levels.  To help illustrate this measure, consider Table \ref{Tbl:PAlevelErrorRatesAndRatiosARthreshPreFC}.  The left-hand side of this table provides, for each level of the protected attribute, the false positive and false negative rates at the average risk threshold for a single random subsample of child-reunification pairs.  Correspondingly, the right-hand side of this table provides, for all possible pairings of protected attribute levels, the pairwise ratio of false negative rates and the pairwise ratio of false positive rates.  

From this table it is clear that neither the false negative rates nor the false positive rates are categorically the same across the levels of the protected attribute.  What remains to be determined, however, is the extent of the observed disproportionality.  To quantify this, we utilize the pairwise error rate ratios.  Consider, for example, the pairwise false positive rate ratio between BL and HPA child-reunification pairs that is provided in row 2, column 1 of the right-hand matrix of Table \ref{Tbl:PAlevelErrorRatesAndRatiosARthreshPreFC}.  The given value of 0.60 is obtained by taking the smaller false positive rate between the BL and HPA levels and dividing it by the larger of the two false positive rates, i.e., $\frac{0.203}{0.342}\approx0.60$.  This value conveys that the rate at which would-be-successful HPA child-reunification pairs are incorrectly predicted to ``fail'' is only 0.60 the rate at which would-be-successful BL child-reunification pairs are incorrectly predicted to ``fail''.  As yet another example,  consider the pairwise false negative rate ratio between NV and WH child-reunification pairs that is provided in row 3, column 4 of the right-hand matrix of Table \ref{Tbl:PAlevelErrorRatesAndRatiosARthreshPreFC}.  The given value of 0.80 is obtained by taking the smaller false negative rate between the NV and WH levels and dividing it by the larger of the two false negative rates, i.e., $\frac{0.309}{0.386}\approx0.80$.  This value conveys that the rate at which would-be-unsuccessful NV child-reunification pairs are incorrectly predicted to ``succeed'' is only 0.80 the rate at which would-be-unsuccessful WH child-reunification pairs are incorrectly predicted to ``succeed''.   These two values, along with the other 10 analogously calculated ratios, each individually shed light on the potential unfairness of the algorithm at the average-risk threshold, but all 12 collectively must be considered in performing a comprehensive assessment.

To achieve such an assessment, we summarize these 12 pairwise error rate ratios into a single number.  In choosing this summary measure, recognize that all pairwise error rate ratios will always be between 0 and 1, where a value of 1 means that the two corresponding levels of the protected attribute have the same false negative rate, or the same false positive rate\footnote{This does not mean that the false positive rate is equal to the false negative rate.}, depending on which error rate is being considered.  Furthermore, the smaller the pairwise error rate ratio, the greater the disproportionality in the respective error rates between the two corresponding levels of the protected attribute.  The smallest of these 12 ratios therefore represents the most egregious instance of disproportionality in error rates across the levels of the protected attribute.  Consequently, to quantify with a single number the extent to which Error Rate Balance is achieved, for a single random subsample, at any specified threshold, we utilize the minimum of the twelve possible pairwise error rate ratios; for the subsample yielding Table \ref{Tbl:PAlevelErrorRatesAndRatiosARthreshPreFC}, this value is 0.60.  While summary options other than the minimum are possible, including for example the median, we have ultimately chosen the minimum since meaningfully increasing this value to an ``acceptable'' level will necessarily ensure that the 11 other pairwise ratios have also reached such a level.  Hence, the procedure we utilize to mitigate a lack of Error Rate Balance seeks to increase this measure towards one.

\begin{table}
  \caption{For each level of the protected attribute, for a single random subsample of child-reunification pairs, the false positive and false negative rates at the average risk threshold are provided in the left-hand side of this table.  Correspondingly, in the right-hand side, the pairwise ratio of false negative rates and the pairwise ratio of false positive rates are provided for all possible pairings of protected attribute levels.  Note that these pairwise error rate ratios are always constructed such that the larger error rate is in the denominator and the smaller is in the numerator, thus ensuring that all ratios are between 0 and 1.  To then quantify with a single number the extent to which Error Rate Balance is achieved for a single random subsample at an arbitrary threshold, we utilize the minimum of the 12 pairwise error rate ratios, which corresponds to 0.60 in the table below.}
  \label{Tbl:PAlevelErrorRatesAndRatiosARthreshPreFC}
  \begin{tabular}{r c c  r c c c c} 
    \toprule
    & \multicolumn{2}{c}{\textit{Error Rate}} &  &\multicolumn{4}{c}{\textit{Pairwise Error Rate Ratios}} \\
    Level & FNR    & FPR   &  & BL &  HPA  & NV    & WH  \\ 
    \midrule
    BL    & 0.331  & 0.342 & \multicolumn{1}{c}{BL}  &      & 0.90 & 0.93 & 0.86 \\  
    HPA   & 0.368  & 0.203 & \multicolumn{1}{c}{HPA} & 0.60 &      & 0.84 & 0.95 \\ 
    NV    & 0.309  & 0.310 & \multicolumn{1}{c}{NV}  & 0.91 & 0.66 &      & 0.80 \\  
    WH    & 0.386  & 0.279 & \multicolumn{1}{c}{WH}  & 0.82 & 0.73 & 0.90 &       \\
    \bottomrule
    \end{tabular}
\end{table}

\subsection{Quantifying Alternative Group-Level Definitions of Algorithmic Fairness}
\label{AppSub:QuantifyingAltGroupLevelDefns}
The step-by-step procedure described in Section \ref{SubSec:Procedure}, and thoroughly detailed in Appendix \ref{App:TheFairnessCorrectionProcedureStepByStep} directly applies to eight of the nine group-level definitions of algorithmic fairness discussed in this manuscript.  More specifically, in addition to Error Rate Balance, the procedure can be directly utilized with Statistical Parity, Overall Accuracy Equality, Predictive Parity, Equal Opportunity, Predictive Equality, Conditional Use Accuracy Equality, and Treatment Equality by replacing $\mbox{ERB}(\vect{\theta}_{\cdot,i})$ with an analogous measure dictated by the particular definition of algorithmic fairness.  This amounts to following the approach presented above, in Appendix \ref{App:QuantifyingErrorRateBalance}, and replacing the False Negative and False Positive Rate with the one or two predictive performance measures dictated by the specified definition of algorithmic fairness.  For example, with Predictive Parity, only the PPV is calculated at each threshold and therefore only the pairwise ratios, across the protected attribute levels, of the PPV are constructed, with the minimum such ratio measuring the extent to which Predictive Parity is achieved.  As yet another example, with Conditional Use Accuracy Equality, both the PPV and NPV are calculated at each threshold.  Hence, all pairwise ratios of PPV's and NPV's are constructed across the levels of the protected attribute, with the minimum of these values representing the extent to which Conditional Use Accuracy Equality is achieved.

\section{Identifying a Fairness "Correction" Procedure}
\label{App:IdentifyingFairnessCorrectionProcedure}
Following the structure and content provided in \cite{BHJKR2017}, we discuss below the three main classes of procedures for mitigating a lack of algorithmic fairness within a binary classification task, as well as a few of the specific approaches.  Where appropriate, we provide our concerns surrounding methods we explored but ultimately elected not to utilize.  In the end, we settled on employing a protected-attribute-level-specific threshold adjustment approach similar to that described in \cite{HPS2016} and \cite{LCM2017}, which lies within the class of post-processing procedures.  This approach is described in Section \ref{SubSec:Procedure} and thoroughly detailed in Appendix \ref{App:TheFairnessCorrectionProcedureStepByStep}.

Before overviewing these methods, however, we want to emphasize two things.  First, there is no ``perfect'' approach to mitigating a lack of algorithmic fairness.  All methods will bring to the surface the tension that exists between accuracy and fairness.  An ``optimal'' fairness-correction procedures is therefore one which maximally increases fairness with minimal cost to accuracy.  

Second, there are several characteristics we currently view as essential in a fairness-correction procedure, though they may come at varying costs to optimality.  In particular, these characteristics include transparency, interpretability, and operationalizability.  We value a procedure that operates outside of the ``black-box'' of the algorithm so that there is no ``mystery'' behind the algorithmic fairness process.  Such transparency allows stakeholders to straightforwardly recognize how unfairness is being mitigated.  When it comes to interpretability, we value a clear answer to the following question: How much did the Error Rate Balance increase\footnote{Additionally, we want to know how much alternative group-levels definitions of fairness changed.}, and corresponding predictive performance measures decrease, at each threshold?  Such interpretability enables stakeholders and decision-makers to clearly recognize the impact of the procedure and to begin to assess the associated trade-offs.  Finally, with respect to operationalizability, we value procedures that are flexible and transferable across use-cases and classifier types.  More specifically, given that we are 1) building multiple machine learning algorithms throughout the life of a Child Welfare case\footnote{See ORRAI's public-facing website for more information: \url{https://www.oregon.gov/DHS/ORRAI/Pages/index.aspx}.} and 2) our expectation is that newer and better classification algorithms will continue to emerge in the future, a procedure that seamlessly operationalizes across both new use-cases and new learners is of immense value.  Collectively, these characteristics represent the lens through which explored procedures were filtered. 

\subsection{An Overview of  ``Correction'' Procedures}
\label{AppSub:AnOverviewOfCorrectionProcedures}
Most approaches to mitigating a lack of algorithmic fairness seem to fall into three identified classes: pre-processing, in-processing, and post-processing.   Pre-processing methods aspire to remove any dependencies between the protected attribute and the other features before training the predictive algorithm.  In other words, such methods seek to prohibit the protected attribute from informing the prediction in any manner, either directly through the inclusion of the protected attribute in the feature set, or indirectly through the protected attribute's association with other features.  Some proposed means for accomplishing this objective include 1) using a sequence of regression models to extract protected attribute information from all other features and 2) rebalancing the prevalence of the outcome variable across the protected attribute levels, potentially via random ``recoding'' \cite{BHJKR2017}.  

Our reasons for dismissing these pre-processing approaches are rooted both in concerns specific to each approach, as well as a concern shared across approaches.  In particular, a rebalancing-based approach ``...implies using different false positive to false negative rates for different protected [attribute levels]'' \cite[][pg. 26]{BHJKR2017}.  In other words, such an approach works in direct opposition to Error Rate Balance.  Any sequential regression approach, on the other hand, either 1) requires anticipating all two-way and higher-order interaction effects across the 500-plus features, which is unrealistic or 2) utilizing a ``...far more sophisticated residualizing process'', which we anticipate would compromise the operationalizability of the approach \cite[][pg. 26]{BHJKR2017}.  Finally, such approaches do not easily lend themselves to exploring the trade-off continuum between fairness and predictive performance.  In particular, it is difficult to identify a clear mechanism within these procedures that enables a clean and informative empirical exploration of the functional relationship between accuracy and fairness. 

The second class of methods, in-processing, aspires to improve fairness by incorporating the corresponding definition into the optimization process of the classification algorithm.  In other words, such methods use the feature set as is, without removing any dependencies between the protected attribute and the other features, but penalize the algorithm if the subsequent predictions violate the specified definition of algorithmic fairness.  Through such a penalty the algorithm learns in a manner that ``values'' both predictive performance and fairness.  An example of such an approach is provided in \cite{ZVRG2017}, where a fairness constraint is embedded within the objective function of both a logistic regression model and a support vector machine.  Our general avoidance of such in-processing methods is solely based on their inability to meet two of the aforementioned essential characteristics.  In particular, embedding the fairness component within the optimizer of the classifier, in the form of a penalty term, precludes the type of transparency we currently desire.  Furthermore, such approaches are not easily operationalized since any time a newer or better learner emerges in the literature, we would have to ``crack open'' the corresponding algorithm and figure out how to efficiently embed the specified fairness constraint.

The remaining class of methods, post-processing, aspires to improve fairness by adjusting predictions after training the algorithm.  In other words, such methods 1) use the feature set as is, without removing any dependencies between the protected attribute and the other features, 2) train the algorithm in a ``traditional'' sense, without any constraint surrounding fairness,  and then 3) ``correct'' the subsequent predictions in accordance with the specified definition of fairness.  One general approach to achieving this is to identify and utilize protected-attribute-level-specific thresholds to mitigate a lack of algorithmic fairness \cite{HPS2016, LCM2017}.   In the context of the reunification algorithm, such an approach means that the amount of evidence\footnote{As provided by the predicted probability, or some ordered analogue, of the classifier.} required to elevate the level of risk\footnote{For example, from a score of 2 to 3.} for a child-reunification pair depends on the level of the protected attribute; furthermore, the ``dial'' for determining the ``appropriate'' amount of evidence for each protected attribute level is ``tuned'' according to the specified definition of algorithmic fairness.  In \cite{LCM2017}, this definition is Statistical Parity\footnote{Technically, the p-percent rule is used as the ``tuning'' mechanism, but the p-percent rule is just a measure quantifying the extent to which Statistical Parity is achieved.}, while in \cite{HPS2016} this definition is Error Rate Balance\footnote{Technically, the procedure is discussed both for Equal Opportunity and Equalized Odds, but Error Rate Balance is equivalent to Equalized Odds.}.  In theory, however, such a procedure can also be applied to any other group-level definition of algorithmic fairness, though potentially sub-optimally. 

The viability of this protected-attribute-level-specific thresholds approach resides in its alignment with our transparency, interpretability, and operationalizability demands.  In particular, the procedure not only occurs outside of the ``black-box'' of the training algorithm, but the high-level how of the approach is also straight-forward to explain.  Additionally, given that the slew of predictive performance and algorithmic fairness\footnote{Including, but not limited to, the one specified.} measures can easily be calculated for both the group-agnostic (i.e., ``pre-fairness-corrected'') threshold values and the group-specific (i.e., ``post-fairness-corrected'') threshold values, determining  how much the Error Rate Balance increased\footnote{Additionally, the extent to which the other group-level algorithmic fairness measures changed can be easily calculated.}at each threshold, as well as how much the corresponding predictive performance measures decreased, is also relatively easy.  Finally, because this type of approach is agnostic to both the use-case and the utilized learner\footnote{So long as the prediction outputted by the learner is quantitative with larger values indicating greater risk of the adverse event (e.g., predicted probabilities).}, operationalizing it is also relatively straight-forward.  

It is worth addressing here a common and instinctive rhetorical question to such a process: But is it not unfair to uphold different ``standards'' for different protected attribute levels when assigning risk scores?  Such a response is, at its core, pushing back against the notion that to prevent disparate impact requires disparate treatment, and is in fact the motivation behind the in-processing approach proposed in \cite{ZVRG2017}.  In reality, however, such approaches seeking to prevent disparate impact without disparate treatment 1) fail to optimally prevent disparate impact and 2) ultimately do enact disparate treatment ``...through hidden changes to the learning algorithm'' \cite[][pg. 16]{LCM2017}.  Hence, such a criticism is not limited to a protected-attribute-level-specific thresholds approach, but in fact applies to a broad range of fairness correction procedures.  And in response to this more general criticism, we point out that such disparate treatment is in fact enacting a core value of the Oregon Department of Human Services -- service equity.

\section{Step-by-Step Details of the Algorithmic Fairness Correction Procedure}
\label{App:TheFairnessCorrectionProcedureStepByStep}
In this section, we provide the step-by-step details of the proposed process for obtaining post-fairness-corrected threshold values for the reunification algorithm.  While the procedure is exemplified for three thresholds -- low-risk, average-risk, and high-risk -- and four protected attribute levels -- BL, HPA, NV, and WH -- the procedure can straightforwardly be generalized to accommodate an arbitrary number of thresholds and protected attribute levels.  The rationale behind the constraints embedded in the objective function of steps 6-8 below is provided in Appendix \ref{AppSub:UnderstandingTheLogicOfTheProcedure}.  

Let $X$ be the $4 \times N$ matrix in which the $N$ rows represent the entire sample of child-reunification pairs and the 4 columns represent, respectively, a unique child identification number (for subsampling purposes), the observed value of the binary outcome variable, the level of the protected attribute, and the predicted probability obtained from the algorithm.
\begin{enumerate}
\item Set number of subsamples, $I$ (e.g., $I=200$).
\item Set the value of the tuning parameter, $w$, where $0\leq w\leq1$.
\item Initialize the subsample index: $i=1$.
\item Obtain the $i^{th}$ random subsample of $\vect{X}$, denoted by $\vect{X}_i$. 
\item Calculate the group-agnostic low-risk, average-risk, and high-risk threshold values, $\Phi_{L,i}, \Phi_{A,i}, \Phi_{H,i}$, corresponding to subsample $i$, where 
	\begin{itemize}
	\item $\Phi_{A,i}$ represents the average predicted probability for the child-reunification pairs of subsample $i$, 
	\item $\Phi_{L,i}$ represents the $50^{th}$ percentile of predicted probabilities for the child-reunification pairs of subsample $i$ that are less than  $\Phi_{A,i}$, and
	\item $\Phi_{H,i}$ represents the $75^{th}$ percentile of predicted probabilities for the child-reunification pairs of subsample $i$ that are greater than  $\Phi_{A,i}$. 
	\item[$\dagger$] Recall that the identification of such cutoffs is rooted in the use-case.
	\end{itemize}
\item Obtain the group-specific low-risk threshold values for subsample $i$, $\widehat{\vect{\theta}_{L,i}} = \bigg(\hat\theta_{L,i}^{BL}, \hat\theta_{L,i}^{HPA}, \hat\theta_{L,i}^{NV}, \hat\theta_{L,i}^{WH}\bigg)'$, by solving the following penalized optimization problem
	\begin{itemize}
	\item[] \begin{argmini}
			{\vect{\theta}_{L,i}}{(1-w)(1-\mbox{ERB}(\vect{\theta}_{L,i})) \;+\; w\Delta(\vect{\theta}_{L,i}, \Phi_{L,i})}     
			{}{}
			\addConstraint{0<}{\min\bigg\{\theta_{L,i}^{BL}, \theta_{L,i}^{HPA}, \theta_{L,i}^{NV}, \theta_{L,i}^{WH}\bigg\}}{ \leq \Phi_{L,i}}
			\addConstraint{\Phi_{L,i}\leq}{\max\bigg\{\theta_{L,i}^{BL}, \theta_{L,i}^{HPA}, \theta_{L,i}^{NV}, \theta_{L,i}^{WH}\bigg\}}{< \Phi_{A,i}},
		\end{argmini}
		where $\mbox{ERB}(\vect{\theta}_{L,i})$ is the value quantifying the extent to which Error Rate Balance is achieved, as defined in Appendix \ref{App:QuantifyingErrorRateBalance}, at $\vect{\theta}_{L,i}$, and $\Delta(\vect{\theta}_{L,i}, \Phi_{L,i})$ is the proportion of risk scores that change (i.e., either from 1 to 2 or from 2 to 1) when moving from the group-agnostic threshold value, $\Phi_{L,i}$, to the group-specific threshold values specified by $\vect{\theta}_{L,i}$.
	\end{itemize}
\item Obtain the group-specific average-risk threshold values for subsample $i$, $\widehat{\vect{\theta}_{A,i}} = \bigg(\hat\theta_{A,i}^{BL}, \hat\theta_{A,i}^{HPA}, \hat\theta_{A,i}^{NV}, \hat\theta_{A,i}^{WH}\bigg)'$, by solving the following penalized optimization problem
	\begin{itemize}
	\item[] \begin{argmini}
			{\vect{\theta}_{A,i}}{(1-w)(1-\mbox{ERB}(\vect{\theta}_{A,i})) \;+\; w\Delta(\vect{\theta}_{A,i}, \Phi_{A,i})} 
			{}{}
			\addConstraint{\hat\theta_{L,i}^{BL}<}{\;\;\;\;\theta_{A,i}^{BL}}{ \leq \Phi_{H,i}}
			\addConstraint{\hat\theta_{L,i}^{HPA}<}{\;\;\;\;\theta_{A,i}^{HPA}}{ \leq \Phi_{H,i}}
			\addConstraint{\hat\theta_{L,i}^{NV}<}{\;\;\;\;\theta_{A,i}^{NV}}{ \leq \Phi_{H,i}}
			\addConstraint{\hat\theta_{L,i}^{WH}<}{\;\;\;\;\theta_{A,i}^{WH}}{ \leq \Phi_{H,i}}
			\addConstraint{\min\bigg\{\theta_{A,i}^{BL}, \theta_{A,i}^{HPA}, \theta_{A,i}^{NV}, \theta_{A,i}^{WH}\bigg\}}{ \leq \Phi_{A,i}}{}
			\addConstraint{\max\bigg\{\theta_{A,i}^{BL}, \theta_{A,i}^{HPA}, \theta_{A,i}^{NV}, \theta_{A,i}^{WH}\bigg\}}{\geq \Phi_{A,i}}{},
		\end{argmini}
		where $\mbox{ERB}(\vect{\theta}_{A,i})$ is the value quantifying the extent to which Error Rate Balance is achieved, as defined in Appendix \ref{App:QuantifyingErrorRateBalance}, at $\vect{\theta}_{A,i}$, and $\Delta(\vect{\theta}_{A,i}, \Phi_{A,i})$ is the proportion of risk scores that change (i.e., either from 2 to 3 or from 3 to 2) when moving from the group-agnostic threshold value, $\Phi_{A,i}$, to the group-specific threshold values specified by $\vect{\theta}_{A,i}$.
	\end{itemize}
\item Obtain the group-specific high-risk threshold values for subsample $i$, $\widehat{\vect{\theta}_{H,i}} = \bigg(\hat\theta_{H,i}^{BL}, \hat\theta_{H,i}^{HPA}, \hat\theta_{H,i}^{NV}, \hat\theta_{H,i}^{WH}\bigg)'$, by solving the following penalized optimization problem
	\begin{itemize}
	\item[] \begin{argmini}
			{\vect{\theta}_{H,i}}{(1-w)(1-\mbox{ERB}(\vect{\theta}_{H,i})) \;+\; w\Delta(\vect{\theta}_{H,i}, \Phi_{H,i})} 
			{}{}
			\addConstraint{\hat\theta_{A,i}^{BL}<}{\;\;\;\;\theta_{H,i}^{BL}}{ < 1}
			\addConstraint{\hat\theta_{A,i}^{HPA}<}{\;\;\;\;\theta_{H,i}^{HPA}}{ < 1}
			\addConstraint{\hat\theta_{A,i}^{NV}<}{\;\;\;\;\theta_{H,i}^{NV}}{ < 1}
			\addConstraint{\hat\theta_{A,i}^{WH}<}{\;\;\;\;\theta_{H,i}^{WH}}{ < 1}
			\addConstraint{\min\bigg\{\theta_{H,i}^{BL}, \theta_{H,i}^{HPA}, \theta_{H,i}^{NV}, \theta_{H,i}^{WH}\bigg\}}{ \leq \Phi_{H,i}}{}
			\addConstraint{\max\bigg\{\theta_{H,i}^{BL}, \theta_{H,i}^{HPA}, \theta_{H,i}^{NV}, \theta_{H,i}^{WH}\bigg\}}{\geq \Phi_{H,i}}{},
		\end{argmini}
		where $\mbox{ERB}(\vect{\theta}_{H,i})$ is the value quantifying the extent to which Error Rate Balance is achieved, as defined in Appendix \ref{App:QuantifyingErrorRateBalance}, at $\vect{\theta}_{H,i}$, and $\Delta(\vect{\theta}_{H,i}, \Phi_{H,i})$ is the proportion of risk scores that change (i.e., either from 3 to 4 or from 4 to 3) when moving from the group-agnostic threshold value, $\Phi_{H,i}$, to the group-specific threshold values specified by $\vect{\theta}_{H,i}$.
	\end{itemize}
\item Increase subsample index: $i = i+1$.  If $i \leq I$, repeat steps 4-8, else move on to step 10.
\item For the specified value of $w$, obtain the pre-fairness-corrected values for the low-risk threshold, $\Phi_{L,w}$, the average-risk threshold, $\Phi_{A,w}$, and the high-risk threshold, $\Phi_{H,w}$.  These values are obtained via the following bagging-like process:
	\begin{itemize}
	\item  $\Phi_{L,w} = \frac{1}{I}\sum_{i=1}^{I}\Phi_{L,i}$, with  $\Phi_{A,w}$ and  $\Phi_{H,w}$ analogously calculated.
	\end{itemize}	
\item For the specified value of $w$, obtain the post-fairness-corrected values for the group-specific low-risk thresholds, $\widehat{\vect{\theta}_{L,w}} = \bigg(\hat\theta_{L,w}^{BL}, \hat\theta_{L,w}^{HPA}, \hat\theta_{L,w}^{NV}, \hat\theta_{L,w}^{WH}\bigg)'$, the group-specific average-risk thresholds, $\widehat{\vect{\theta}_{A,w}} = \bigg(\hat\theta_{A,w}^{BL}, \hat\theta_{A,w}^{HPA}, \hat\theta_{A,w}^{NV}, \hat\theta_{A,w}^{WH}\bigg)'$, and the group-specific high-risk thresholds, \newline $\widehat{\vect{\theta}_{H,w}} = \bigg(\hat\theta_{H,w}^{BL}, \hat\theta_{H,w}^{HPA}, \hat\theta_{H,w}^{NV}, \hat\theta_{H,w}^{WH}\bigg)'$.  These values are obtained via the following bagging-like process:
 	\begin{itemize}
	\item  $\hat\theta_{L,w}^{BL} = \frac{1}{I}\sum_{i=1}^{I}\hat\theta_{L,i}^{BL}$, with  $\hat\theta_{L,w}^{HPA}$, $\hat\theta_{L,w}^{NV}$, and $\hat\theta_{L,w}^{WH}$ similarly calculated.  The constituent components of  $\widehat{\vect{\theta}_{A,w}}$ and $\widehat{\vect{\theta}_{H,w}}$ are analogously calculated as well.  
	\end{itemize}
\end{enumerate}
\subsection{Understanding the Constraints of the Optimization Procedure}
\label{AppSub:UnderstandingTheLogicOfTheProcedure}
This section clarifies the objectives of the constraints associated with the optimization problem in steps 6-8 of Appendix \ref{App:TheFairnessCorrectionProcedureStepByStep}.  In particular, through these constraints the identified group-specific threshold values are guaranteed to exist (i.e., be between 0 and 1) and to exhibit two properties we label as ``orderliness''  and ``coveredness''.  By orderliness, we mean that within a particular level of the protected attribute, it must be the case that the low-risk threshold value is less than the average-risk threshold value, which is in turn less than the high-risk threshold value (e.g., $\hat\theta_{L,i}^{BL}  < \hat\theta_{A,i}^{BL} < \hat\theta_{H,i}^{BL}$).  By coveredness, we mean that at a particular threshold, the group-agnostic threshold value must be within the closed interval created by the corresponding minimum and maximum group-specific threshold values (e.g., $\Phi_{A,i} \in [\min{\widehat{\vect{\theta}_{A,i}}}, \max{\widehat{\vect{\theta}_{A,i}}}]$).  The logic with this coveredness property is to retain, as much as possible, the meaning and intent behind the ``original'', group-agnostic, threshold values that were identified as part of the business case motivating the algorithm's development. 
\section{Simultaneous Penalized Optimization for Calibration}
\label{App:SimultaneousPenalizedOptimizationCalibration}

The step-by-step procedure described in Section \ref{SubSec:Procedure}, and thoroughly detailed in Appendix \ref{App:TheFairnessCorrectionProcedureStepByStep} directly applies to eight of the nine group-level definitions of algorithmic fairness discussed in this manuscript.  More specifically, in addition to Error Rate Balance, the procedure can be directly utilized with Statistical Parity, Overall Accuracy Equality, Predictive Parity, Equal Opportunity, Predictive Equality, Conditional Use Accuracy Equality, and Treatment Equality by replacing $\mbox{ERB}(\vect{\theta}_{\cdot,i})$ with an analogous measure dictated by the particular definition of algorithmic fairness; see Appendix \ref{AppSub:QuantifyingAltGroupLevelDefns} for the details of such replacement.

For Calibration, however, the sequential penalized optimization problems in steps 6-8 of Appendix \ref{App:TheFairnessCorrectionProcedureStepByStep} must be combined into one simultaneous penalized optimization problem, given below.  Additionally, note that $\mbox{ERB}(\vect{\theta}_{\cdot,i})$, which measures the extent two which Error Rate Balance is achieved at an arbitrary threshold for subsample $i$ (so $\vect{\theta}_{\cdot,i}=\vect{\theta}_{L,i},\;\vect{\theta}_{A,i}$, or $\vect{\theta}_{H,i}$), is now replaced with $\mbox{CAL}(S_{i},\vect{\theta}_{T,i})$, which measures the extent to which Calibration is achieved at risk score $S\in\{S1,S2,S3,S4\}$ for subsample $i$, which depends on one or more thresholds in the set of all thresholds (so $\vect{\theta}_{T,i} = (\vect{\theta}_{L,i},\; \vect{\theta}_{A,i},\; \vect{\theta}_{H,i})'$).  Hence, to calculate $\mbox{CAL}(S_{i},\vect{\theta}_{T,i})$ at each risk score $S\in\{S1,S2,S3,S4\}$, the approach presented in Appendix \ref{App:QuantifyingErrorRateBalance} can analogously be used, though it is calculated at each risk score rather than at each threshold.  More specifically, the process amounts to first calculating the proportion of observational units assigned risk score S, where $S\in\{1,2,3,4\}$ that go on to experience the adverse event, and then calculating the pairwise ratios of such a proportion across each pairing of protected attribute levels; the minimum such ratio represents the extent to which Calibration is achieved at a particular risk score.  Consequently, Calibration can be utilized within the developed fairness correction procedure by replacing steps 6-8 of Appendix \ref{App:TheFairnessCorrectionProcedureStepByStep} with the following step:
\begin{itemize}
\item Obtain the group-specific threshold values across the low-, average-, and high-risk thresholds, for the $i^{th}$ subsample, $\widehat{\vect{\theta}_{T,i}} = \bigg(\widehat{\vect{\theta}_{L,i}},\; \widehat{\vect{\theta}_{A,i}},\; \widehat{\vect{\theta}_{H,i}}\bigg)'$, where
\newline $\widehat{\vect{\theta}_{L,i}} = \bigg(\hat\theta_{L,i}^{BL},\; \hat\theta_{L,i}^{HPA},\; \hat\theta_{L,i}^{NV},\; \hat\theta_{L,i}^{WH}\bigg)'$, \newline $\widehat{\vect{\theta}_{A,i}} = \bigg(\hat\theta_{A,i}^{BL},\; \hat\theta_{A,i}^{HPA},\; \hat\theta_{A,i}^{NV},\; \hat\theta_{A,i}^{WH}\bigg)'$, and \newline $\widehat{\vect{\theta}_{H,i}} = \bigg(\hat\theta_{H,i}^{BL},\; \hat\theta_{H,i}^{HPA},\; \hat\theta_{H,i}^{NV},\; \hat\theta_{H,i}^{WH}\bigg)'$, by solving the following penalized optimization problem
	\begin{itemize}
	\item[] \begin{argmini}
			{\vect{\theta}_{T,i}}{w\Delta(\vect{\theta}_{T,i}, \vect{\Phi_{T,i}}) \;+\; (1-w)\sum_{S\in\{S1,S2,S3,S4\}}(1-\mbox{CAL}(S_{i},\vect{\theta}_{T,i}))}     
			{}{}
			\addConstraint{}{\min\bigg\{\theta_{L,i}^{BL}, \theta_{L,i}^{HPA}, \theta_{L,i}^{NV}, \theta_{L,i}^{WH}\bigg\}}{\leq \Phi_{L,i}}
			\addConstraint{}{\max\bigg\{\theta_{L,i}^{BL}, \theta_{L,i}^{HPA}, \theta_{L,i}^{NV}, \theta_{L,i}^{WH}\bigg\}}{\geq \Phi_{L,i}}
			\addConstraint{}{\min\bigg\{\theta_{A,i}^{BL}, \theta_{A,i}^{HPA}, \theta_{A,i}^{NV}, \theta_{A,i}^{WH}\bigg\}}{ \leq \Phi_{A,i}}
			\addConstraint{}{\max\bigg\{\theta_{A,i}^{BL}, \theta_{A,i}^{HPA}, \theta_{A,i}^{NV}, \theta_{A,i}^{WH}\bigg\}}{\geq \Phi_{A,i}}
			\addConstraint{}{\min\bigg\{\theta_{H,i}^{BL}, \theta_{H,i}^{HPA}, \theta_{H,i}^{NV}, \theta_{H,i}^{WH}\bigg\}}{ \leq \Phi_{H,i}}
			\addConstraint{}{\max\bigg\{\theta_{H,i}^{BL}, \theta_{H,i}^{HPA}, \theta_{H,i}^{NV}, \theta_{H,i}^{WH}\bigg\}}{\geq \Phi_{H,i}}
			\addConstraint{0<}{\;\;\;\;\;\;\theta_{L,i}^{BL}\;\;<\;\;\;\;\;\theta_{A,i}^{BL}\;\;<\;\;\theta_{H,i}^{BL}}{<1}
			\addConstraint{0<}{\;\;\;\theta_{L,i}^{HPA}\;\;<\;\;\theta_{A,i}^{HPA}\;\;<\;\;\theta_{H,i}^{HPA}}{<1}
			\addConstraint{0<}{\;\;\;\;\;\theta_{L,i}^{NV}\;\;<\;\;\;\;\theta_{A,i}^{NV}\;\;<\;\;\theta_{H,i}^{NV}}{<1}
			\addConstraint{0<}{\;\;\;\;\theta_{L,i}^{WH}\;\;<\;\;\;\theta_{A,i}^{WH}\;\;<\;\;\theta_{H,i}^{WH}}{<1},
		\end{argmini}
	\end{itemize}
	where $\vect{\Phi_{T,i}}=(\Phi_{L,i},\;\Phi_{A,i}\;\Phi_{H,i})'$ and $\Delta(\vect{\theta}_{T,i}, \vect{\Phi_{T,i}})$ is the proportion of risk scores that change in transitioning from the group-agnostic to group-specific threshold values.
\end{itemize}
\section{Demonstrating Generalizability}
\label{App:DemonstratingGeneralizability}
To demonstrate the broad applicability of the presented fairness-correction procedure, we applied it, as detailed in Appendix \ref{App:TheFairnessCorrectionProcedureStepByStep}, under three alternative definitions of algorithmic fairness, including Conditional Use Accuracy Equality, Treatment Equality, and Calibration\footnote{The details of the optimization problem for this definition of algorithm fairness are detailed in Appendix \ref{App:SimultaneousPenalizedOptimizationCalibration}}.  In each scenario, 200 random subsamples were utilized (i.e., $I=200$) and a grid of 101 $w$-values ranging between zero and one were considered (i.e., $w=0, 0.01,\ldots, 0.99,1$).  For each of these three scenarios, for all 101 $w$-values, the procedure detailed in Appendix \ref{App:IdentifyingBestPostFairnessCorrectedThresholdValuesStepByStep} was then utilized to determine the extent to which the specified definition of fairness was achieved, on average, across 200 random subsamples (i.e., $J=200$). For each of the three scenarios, the "best" $w$-values across the three thresholds were then identified, where "best" corresponds to the post-fairness-corrected threshold values producing the greatest extent to which the specific definition of algorithmic fairness was achieved, without any regard for accuracy costs.  Condensed results for each scenario are provided below.

\subsection{Results Under Conditional Use Accuracy Equality (CUAE)}
The "best" $w$-values in this scenario are $w=0.00, 0.01, 0.02$ for the low-risk, average-risk, and high-risk thresholds respectively.  The procedure detailed in Appendix \ref{App:IdentifyingBestPostFairnessCorrectedThresholdValuesStepByStep} was run one additional time across 200 random subsamples (i.e., $J=200$) with these threshold-specific $w$-values.  Across these 200 subsamples, the average number of risk scores that changed in transitioning from the pre- to post-fairness-corrected threshold values was 10.70\%.  The average extent to which Conditional Use Accuracy Equality was achieved at each threshold, both for the pre- and post-fairness-corrected threshold values, is provided in Table \ref{Tbl:CUAEresults}, along with the corresponding standard deviations.
\begin{table*}[t]
    \caption{The average extent to which Conditional Use Accuracy Equality was achieved at each threshold, both for the pre- and post-fairness-corrected threshold values, is provided, along with the corresponding standard deviations.}
    \label{Tbl:CUAEresults}
    \begin{tabular}{c|cc|cc}
    \toprule
    & \multicolumn{2}{c|}{\textbf{Pre-Fairness-Corrected}} & \multicolumn{2}{c}{\textbf{Post-Fairness-Corrected}}\\
    Threshold & $\overline{\mbox{CUAE}}(\Phi_{\cdot, w}, \vect{X})$ & $\text{SD}(\mbox{CUAE}(\Phi_{\cdot, w}, \vect{X}))$ & $\overline{\mbox{CUAE}}(\widehat{\vect{\theta}_{\cdot, w}}, \vect{X})$ & $\text{SD}(\mbox{CUAE}(\widehat{\vect{\theta}_{\cdot, w}}, \vect{X}))$ \\ 
    \midrule
    Low-Risk  & 0.84 & 0.03 & 0.95 & 0.02  \\
    Average-Risk & 0.90 & 0.02 & 0.94 & 0.02 \\
    High-Risk & 0.82 & 0.02 & 0.93 & 0.02 \\
    \bottomrule
    \end{tabular}
\end{table*}
From this table, it is evident that the procedure is successful at increasing CUAE at all three thresholds.

\subsection{Results Under Treatment Equality (TE)}
The "best" $w$-values in this scenario are $w=0.84, 0.74, 0.90$ for the low-risk, average-risk, and high-risk thresholds respectively.  The procedure detailed in Appendix \ref{App:IdentifyingBestPostFairnessCorrectedThresholdValuesStepByStep} was run one additional time across 200 random subsamples (i.e., $J=200$) with these threshold-specific $w$-values.  Across these 200 subsamples, the average number of risk scores that changed in transitioning from the pre- to post-fairness-corrected threshold values was 1.42\%.  The average extent to which Treatment Equality was achieved at each threshold, both for the pre- and post-fairness-corrected threshold values, is provided in Table \ref{Tbl:TEresults}, along with the corresponding standard deviations.
\begin{table*}[t]
    \caption{The average extent to which Treatment Equality was achieved at each threshold, both for the pre- and post-fairness-corrected threshold values, is provided, along with the corresponding standard deviations.}
    \label{Tbl:TEresults}
    \begin{tabular}{c|cc|cc}
    \toprule
    & \multicolumn{2}{c|}{\textbf{Pre-Fairness-Corrected}} & \multicolumn{2}{c}{\textbf{Post-Fairness-Corrected}}\\
    Threshold & $\overline{\mbox{TE}}(\Phi_{\cdot, w}, \vect{X})$ & $\text{SD}(\mbox{TE}(\Phi_{\cdot, w}, \vect{X}))$ & $\overline{\mbox{TE}}(\widehat{\vect{\theta}_{\cdot, w}}, \vect{X})$ & $\text{SD}(\mbox{TE}(\widehat{\vect{\theta}_{\cdot, w}}, \vect{X}))$ \\ 
    \midrule
    Low-Risk  & 0.79 & 0.07 & 0.86 & 0.06   \\
    Average-Risk & 0.83 & 0.05 & 0.91 & 0.04 \\
    High-Risk & 0.82 & 0.05 & 0.91 & 0.04 \\ \bottomrule
    \end{tabular}
\end{table*}
From this table, it is evident that the procedure is successful at increasing TE at all three thresholds.

\subsection{Results Under Calibration (CAL)}
The "best" $w$-values in this scenario were the same, $w=0.05$, at all three thresholds since a single penalized optimization problem must be solved across all three thresholds simultaneously, as detailed in Appendix \ref{App:SimultaneousPenalizedOptimizationCalibration}.  The procedure detailed in Section \ref{App:IdentifyingBestPostFairnessCorrectedThresholdValuesStepByStep} was then run one additional time across 200 random subsamples (i.e., $J=200$) with $w=0.05$.  Across these 200 subsamples, the average number of risk scores that changed in transitioning from the pre- to post-fairness-corrected threshold values was 5.57\%.  The average extent to which Calibration was achieved at each risk score, both for the pre- and post-fairness-corrected threshold values, is provided in Table \ref{Tbl:CALresults}, along with the corresponding standard deviations.
\begin{table*}[t]
    \caption{The average extent to which Calibration was achieved at each risk score, both for the pre- and post-fairness-corrected threshold values, is provided, along with the corresponding standard deviations.}
    \label{Tbl:CALresults}
    \begin{tabular}{c|cc|cc}
    \toprule
    & \multicolumn{2}{c|}{\textbf{Pre-Fairness-Corrected}} & \multicolumn{2}{c}{\textbf{Post-Fairness-Corrected}}\\
    Threshold & $\overline{\mbox{CAL}}(\Phi_{\cdot, w}, \vect{X})$ & $\text{SD}(\mbox{CAL}(\Phi_{\cdot, w}, \vect{X}))$ & $\overline{\mbox{CAL}}(\widehat{\vect{\theta}_{\cdot, w}}, \vect{X})$ & $\text{SD}(\mbox{CAL}(\widehat{\vect{\theta}_{\cdot, w}}, \vect{X}))$ \\ 
    \midrule
    S1 & 0.53 & 0.06 & 0.71 & 0.07 \\
    S2 & 0.73 & 0.05 & 0.89 & 0.05 \\
    S3 & 0.91 & 0.04 & 0.93 & 0.03 \\
    S4 & 0.82 & 0.02 & 0.84 & 0.02 \\ 
    \bottomrule
    \end{tabular}
\end{table*}
From this table, it is evident that the procedure is successful at increasing CAL across all four risk scores.
\section{Need for Training Versus Testing Splits}
\label{App:NeedTrainingTestingSplitsConvo}
A complicated issue is the potential optimism in the presented results of Section \ref{Sec:ResultsFairnessCorrectionProcedure}.  Because we have not utilized training and testing splits, or some form of cross-validation, there is the likely possibility that these results favorably estimate the effectiveness of the developed procedure.  The limiting factors preventing us from employing either of these established methods are two-fold.  First, in order to estimate with "reasonable" precision the extent to which Error Rate Balance is achieved, across subsamples, a relatively large number of observational units are required across each level of the protected attribute.  Given that we already pushed the bounds of such precision in utilizing a four-level protected attribute, any traditional splitting of the observational units into training and testing sets\footnote{Such as a 2/3 to 1/3 training versus testing split.} would have rendered this precision, and therefore the procedure itself, next to useless.  In such situations it is logical to instead attempt some form of cross-validation with extremely small fold sizes, such as leave-one-out cross validation.  This, however, is where the second limiting factor surfaces.  In particular, the computational expense associated with the procedure, while parallelizable,  is non-trivial.  This means applying the procedure across an extremely large number of folds would currently be prohibitively expensive.

In light of these realities, we are left with potentially optimistic results.  There is, however, some question as to whether such validation is truly even necessary in this use-case.  In particular, as described in Section \ref{SubSec:ModelToTool}, the developed procedure employed for estimating the predicted probabilities utilized in the fairness-correction procedure were each based on an algorithm which did not have access to that observation, or any related observation, during model training.  Hence, the fairness-correction procedure identifies the set of post-fairness-corrected threshold values using predicted probabilities that resemble, as much as possible, what will be outputted by the tool once implemented.  Second, the "sample" set of child-reunification pairs used in the development of the tool actually represents the entire population of such observations.  Hence, the concern of "over-fitting" the identified threshold values to the sample, which is a primary motivation behind methods like cross-validation, feels somewhat misplaced.

\section{Step-By-Step Details for Identifying "Best" Penalty Weights}
\label{App:IdentifyingBestPostFairnessCorrectedThresholdValuesStepByStep}
The step-by-step details for calculating the predictive performance and algorithmic fairness of the pre- and post-fairness corrected threshold values obtained for each $w$-value is provided below.
\begin{enumerate}
\item Set number of subsamples, J (e.g., $J=200$).
\item Initialize the subsample index: $j=1$.
\item Obtain the $j^{th}$ random subsample of $\vect{X}$, denoted by $\vect{X}_j$. 
\item Calculate the extent to which Error Rate Balance is achieved for the $j^{th}$ subsample at the pre-fairness-corrected low-risk threshold value, denoted $\mbox{ERB}(\Phi_{L,w}, \vect{X}_j)$, and at the post-fairness-corrected low-risk threshold values, denoted \newline $\mbox{ERB}(\widehat{\vect{\theta}_{L,w}}, \vect{X}_j)$.  Additionally, perform the analogous calculations at the average-risk and high-risk thresholds.
\item Calculate various performance measures for the $j^{th}$ subsample at the pre-fairness-corrected low-risk threshold value, $\Phi_{L,w}$, and at the post-fairness-corrected low-risk threshold values, $\widehat{\vect{\theta}_{L,w}}$.  These performance measures include the false negative rate, true positive rate, false positive rate, true negative rate, positive predictive value, false discovery rate, negative predictive value, false omission rate, and accuracy for each level of the protected attribute, as well as overall.  Additionally, perform the analogous calculations at the average-risk and high-risk thresholds.
\item Increase subsample index: $j = j+1$.  If $j \leq J$, repeat steps 3-5, else move on to step 7.
\item Obtain the average and standard deviation of extent to which Error Rate Balance is achieved across the J subsamples at both the pre-fairness-corrected value for the low-risk threshold and the post-fairness-corrected values for the low-risk threshold.  Obtain analogous measures for the average-risk and high-risk thresholds.  These values are calculated as follows:
	\begin{itemize}
	\item $\overline{\mbox{ERB}}(\Phi_{L,w}, \vect{X}) = \frac{1}{J}\sum_{j=1}^{J}\mbox{ERB}(\Phi_{L,w}, \vect{X}_j)$.
	\item[] $\overline{\mbox{ERB}}(\Phi_{A,w}, \vect{X})$ and $\overline{\mbox{ERB}}(\Phi_{H,w}, \vect{X})$ are analogously calculated.
	\item $\text{SD}(\mbox{ERB}(\Phi_{L,w}, \vect{X})) =$ \newline $\sqrt{\frac{1}{J-1}\sum_{j=1}^{J}(\mbox{ERB}(\Phi_{L,w}, \vect{X}_j) - \overline{\mbox{ERB}}(\Phi_{L,w}, \vect{X}))^2}$.
	\item[] $\text{SD}(\mbox{ERB}(\Phi_{A,w}, \vect{X}))$ and $\text{SD}(\mbox{ERB}(\Phi_{H,w}, \vect{X}))$ are analogously calculated.
	\item $\overline{\mbox{ERB}}(\widehat{\vect{\theta}_{L,w}}, \vect{X}) = \frac{1}{J}\sum_{j=1}^{J}\mbox{ERB}(\widehat{\vect{\theta}_{L,w}}, \vect{X}_j)$.
	\item[] $\overline{\mbox{ERB}}(\widehat{\vect{\theta}_{A,w}}, \vect{X})$ and $\overline{\mbox{ERB}}(\widehat{\vect{\theta}_{H,w}}, \vect{X})$ are analogously calculated.
	\item $\text{SD}(\mbox{ERB}(\widehat{\vect{\theta}_{L,w}}, \vect{X})) = \newline \sqrt{\frac{1}{J-1}\sum_{j=1}^{J}(\mbox{ERB}(\widehat{\vect{\theta}_{L,w}}, \vect{X}_j) - \overline{\mbox{ERB}}(\widehat{\vect{\theta}_{L,w}}, \vect{X}))^2}$.
	\item[] $\text{SD}(\mbox{ERB}(\widehat{\vect{\theta}_{A,w}}, \vect{X}))$ and $\text{SD}(\mbox{ERB}(\widehat{\vect{\theta}_{H,w}}, \vect{X}))$ are analogously calculated.
	\end{itemize}
\item Obtain the average and standard deviation for the various performance measures of Step 5 across the J subsamples.  Such numerical summaries are calculated in an intuitive manner analogous to the calculations of Step 7.
\end{enumerate} 

\section{Additional Details}
\label{App:AdditionalDetails}
In this section, additional figures, tables, and commentary are provided detailing the impact of the fairness correction procedure on the risk scores of the reunification algorithm. 

\subsection{Visualizing Overall Predictive Performance Measures Versus Penalty Weight}
\label{AppSub:OvrlPPvsPenaltyWeight}
\begin{figure*}[t]
    \centering
    \includegraphics[width=\linewidth, height=2.0in]{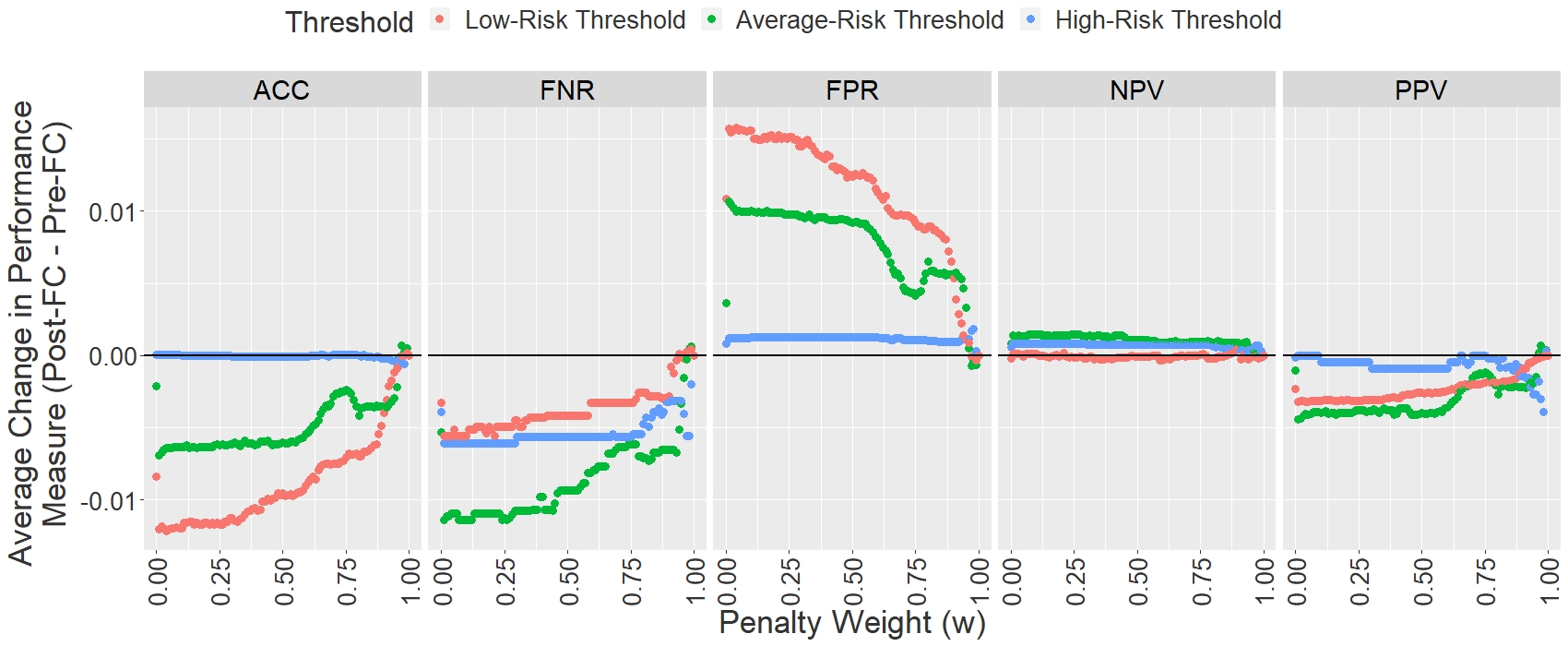}
    \caption{For each of the low-, average-, and high-risk thresholds, the average change in predictive performance (post-fairness correction (Post-FC) value minus pre-fairness correction (Pre-FC) value) versus the penalty weight, $w$, across 200 random subsamples, is plotted for each of five predictive performance measures: Accuracy (ACC), False Negative Rate (FNR), False Positive Rate (FPR), Negative Predictive Value (NPV), and Positive Predictive Value (PPV)}
    \label{Figure:OvrlPPvsPenaltyWeight}
\end{figure*}

For each of the low-, average-, and high-risk thresholds, the average change in predictive performance (post-fairness-corrected (PostFC) value minus pre-fairness-corrected (PreFC) value) versus the penalty weight, $w$, across 200 random subsamples, is plotted in Figure \ref{Figure:OvrlPPvsPenaltyWeight} for each of five predictive performance measures: Accuracy (ACC), False Negative Rate (FNR), False Positive Rate (FPR), Negative Predictive Value (NPV), and Positive Predictive Value (PPV).  From this plot it is evident, for example, that the overall accuracy (i.e., ACC) of the post-fairness-corrected risk scores at the average- and low-risk thresholds tends to be less, at each penalty weight, than that of the pre-fairness-corrected risk scores.  This inferiority in accuracy, however, tends to decrease as $w$ increases towards one, as expected.  Interestingly, at the high-risk threshold, there is essentially no difference in accuracy between the pre- and post-fairness-corrected risk scores.  While similar assessments can be drawn from each of the various predictive performance measures, it worth noting the scale of the vertical axis which ranges from roughly -0.01 to 0.02; in other words, the greatest change in any predictive performance measure, for any penalty weight at any threshold, is approximately 2 percentage points.

\subsection{Visualizing the Thresholds Across a Single Subsample}
\label{AppSub:VisualizingThresholds}
\begin{figure*}[t]
    \centering
    \includegraphics[width=\linewidth, height=2.5in]{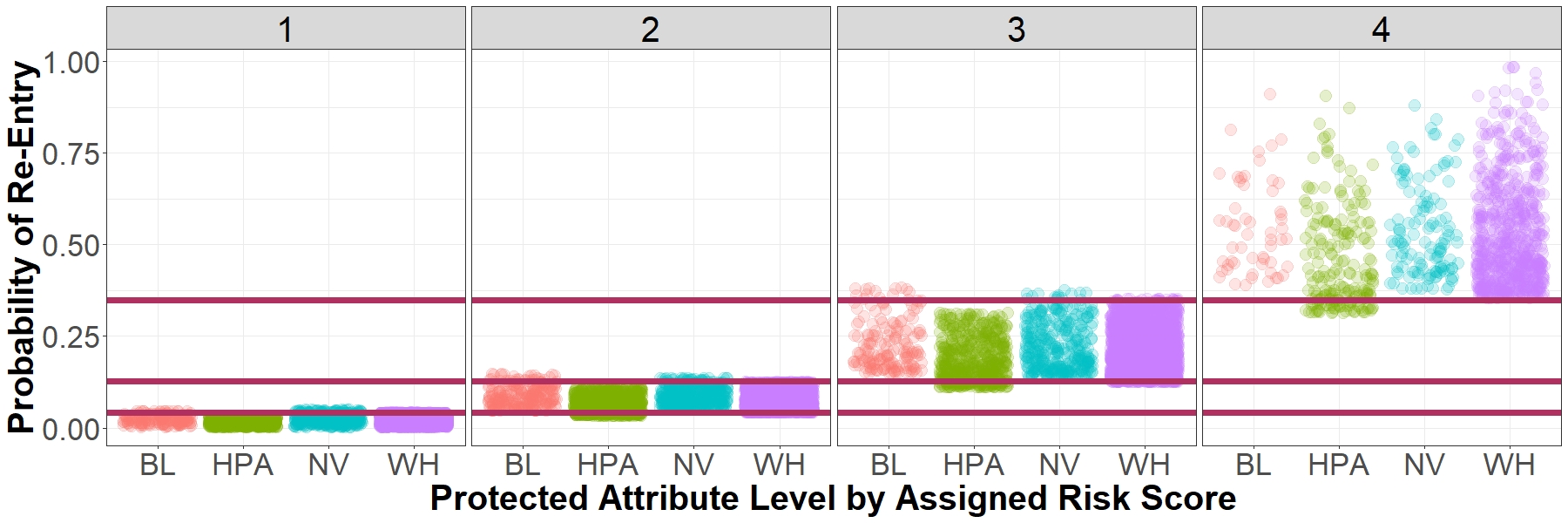}
    \caption{The predicted probability for each child-reunification pair, within a single random subsample, is plotted.  The four panels correspond to the risk scores assigned under the post-fairness-corrected threshold values, grouped according to protected attribute level, while the three horizontal maroon lines running across the panels correspond to the pre-fairness-corrected threshold values.}
    \label{Figure:ImpactPlot}
\end{figure*}

To visualize the change in scores resulting from the fairness-corrected risk scores, consider Figure \ref{Figure:ImpactPlot}, where for a single random subsample, each point represents a predicted probability for a unique child-reunification pair.  The four panels of this figure correspond to the post-fairness-corrected risk scores, grouped according to protected attribute level, and the three horizontal maroon lines correspond to the pre-fairness-corrected threshold values\footnote{Another researcher in ORRAI, Jason Wallin,
recognized the need for such a plot in communicating this work and wrote the R code yielding this graphic.}.  From this plot, the practical impact of the fairness-correction procedure is evident.  For example, in panel 3, a small number of BL child-reunification pairs fall above the upper-most maroon line, indicating that these observational units would have been assigned a risk score of 4 under the pre-fairness-corrected threshold values, but are now assigned a risk score of 3 under the post-fairness-corrected threshold values.  Similarly, in that same panel, a small number of HPA child-reunification pairs fall below the middle maroon line, indicating that these observational units would have been assigned a risk score of 2 under the pre-fairness-corrected threshold values, but are now assigned a risk score of 3 under the post-fairness-corrected threshold values.  

\subsection{Protected-Attribute-Specific False Positive and False Negative Rates}
\label{AppSub:PAspecificFPRsAndFNRs}
Across the $J=200$ random subsamples that were used to obtain the algorithmic fairness and overall predictive performance summary measures provided in Tables \ref{Tbl:ERBresults} and \ref{Tbl:AccuracyResults}, the corresponding average false positive and false negative rates, across all four protected attribute levels, are provided in Table \ref{Tbl:PAspecificFNRsFPRs}.
\begin{table*}[t]
    \caption{The average false negative rate (FNR) and false positive rate (FPR), for both the pre-fairness-corrected (Pre-FC) and post-fairness-corrected (Post-FC) threshold values, across 200 random subsamples.}
    \label{Tbl:PAspecificFNRsFPRs}
    \begin{tabular}{cc|cc|cc}
    \toprule
    && \multicolumn{2}{c|}{\textbf{FNR}} & \multicolumn{2}{c}{\textbf{FPR}} \\
    Threshold & Protected Attribute Level & Pre-FC & Post-FC & Pre-FC & Post-FC \\
    \midrule
    Low-Risk & BL  & 0.113 & 0.129 & 0.722 & 0.697 \\
    Low-Risk & HPA & 0.151 & 0.116 & 0.532 & 0.611 \\
    Low-Risk & NV  & 0.105 & 0.117 & 0.680 & 0.640 \\
    Low-Risk & WH  & 0.131 & 0.128 & 0.634 & 0.642 \\
    \midrule
    Average-Risk & BL  & 0.352 & 0.383 & 0.340 & 0.294  \\
    Average-Risk & HPA & 0.381 & 0.324 & 0.206 & 0.252  \\
    Average-Risk & NV  & 0.325 & 0.343 & 0.310 & 0.287  \\
    Average-Risk & WH  & 0.390 & 0.381 & 0.279 & 0.287 \\
    \midrule
    High-Risk & BL  & 0.706 & 0.755 & 0.057 & 0.047 \\
    High-Risk & HPA & 0.756 & 0.690 & 0.034 & 0.047 \\
    High-Risk & NV  & 0.669 & 0.683 & 0.055 & 0.046 \\
    High-Risk & WH  & 0.769 & 0.771 & 0.051 & 0.050 \\
    \bottomrule
    \end{tabular}
\end{table*}

\section{Child-Transition Features for Machine Learning Classifier}
\label{App:FeaturesForClassifier}
Table \ref{Tbl:FeatureList} lists the features available for each child-transition observation, constructed from administrative data. Features described with bracketed terms (e.g., \# Days) represent multiple which vary over a set of options, such as the number of days into the past, or the type of allegation named in a report of abuse/neglect. These sets of options will vary between jurisdiction. Careful consideration must be made in selecting and defining appropriate features in order to ensure the features represent their intended constructs in a valid and reliable fashion.

\begin{table*}[t]
    \caption{List of child-transition features used in the machine learning classifier.}
    \label{Tbl:FeatureList}
    \begin{tabular}{l|l}
    \toprule
    \textbf{Feature Name} & \textbf{Description} \\
    \midrule
AgeAtStart	&	Current age of child	\\
Gender\_Male	&	Child is male	\\
nuPERPS\_roleVCT	&	\# prior alleged perps associated with the child	\\
nuRPT\_roleVCT\_FM	&	\# prior reports for which child was the victim	\\
nuPERPS\_roleVCT\_FM	&	\# prior alleged perps associated with the child	\\
SP\_n\_\{\# Days\}	&	\# service placements in past \{30,60,90,180,365 Days\}	\\
SP\_n\_\{Placement Type\}\_\{\# Days\}	&	\# service placements in past \{\# Days\} of \{Placement Type Category\}	\\
SP\_n\_InHome\_\{\# Days\}	&	\# In-Home service episodes in past \{\# Days\}	\\
INV\_FM\_n\_\{\# Days\}	&	\# investigations in past \{\# Days\}	\\
INV\_FM\_n\_Safety\_Decision\_\{\# Days\}	&	\# investigations in past \{\# Days\} with Safety Decision of Unsafe	\\
INV\_FM\_n\_Founded\_\{Allegation Type\}\_\{\# Days\}	&	\# founded investigations in past \{\# Days\} with \{Allegation\}	\\
INV\_FM\_n\_UTD\_\{Allegation Type\}\_\{\# Days\}	&	\# undetermined investigations in past \{\# Days\} with \{Allegation\}	\\
INV\_FM\_n\_Unfounded\_\{Allegation Type\}\_\{\# Days\}	&	\# unfounded investigations in past \{\# Days\} with \{Allegation\}	\\
INV\_FM\_n\_NotSDU\_\{Allegation Type\}\_\{\# Days\}	&	\# investigations in past \{\# Days\} with \{Allegation\}	\\
INV\_NONFM\_n\_\{\# Days\}	&	\# non-familial investigations in past \{\# Days\}	\\
RPT\_FM\_n\_Assigned\_\{\# Days\}	&	\# assigned reports involving the child in past \{\# Days\}	\\
RPT\_FM\_n\_\{\# Days\}	&	\# reports involving the child in past \{\# Days\}	\\
RPT\_FM\_n\_Role\_Victim\_\{\# Days\}	&	\# reports involving the child as victrim in past \{\# Days\}	\\
RPT\_FM\_n\_Role\_Perp\_\{\# Days\}	&	\# reports involving the child as perp in past \{\# Days\}	\\
RPT\_FM\_n\_Assigned\_\{Allegation Type\}\_\{\# Days\}	&	\# assigned reports involving the child in past \{\# Days\} with \{Allegation\}	\\
RPT\_FM\_n\_CAS\_\{Allegation Type\}\_\{\# Days\}	&	\# closed reports involving child in past \{\# Days\} with \{Allegation\}	\\
RPT\_NONFM\_n\_\{\# Days\}	&	\# non-familial referrals in past \{\# Days\}	\\
\{Parent\}\_RPT\_asPERP\_n\_Assigned\_\{\# Days\}	&	\# assigned reports involving \{ Parent\} in past \{\# Days\}	\\
\{Parent\}\_RPT\_asPERP\_n\_\{\# Days\}	&	\# reports involving \{ Parent\} in past \{\# Days\}	\\
\{Parent\}\_RPT\_asPERP\_n\_Assigned\_\{Allegation Type\}\_\{\# Days\}	&	\# assigned reports involving \{ Parent\} in past \{\# Days\} with \{Allegation\}	\\
\{Parent\}\_RPT\_asPERP\_n\_CAS\_\{Allegation Type\}\_\{\# Days\}	&	\# reports involving \{ Parent\} in past \{\# Days\}	\\
\{Parent\}\_INV\_asPERP\_n\_\{\# Days\}	&	\# investigations involving \{ Parent\} in past \{\# Days\}	\\
\{Parent\}\_INV\_asPERP\_n\_Founded\_\{Allegation Type\}\_\{\# Days\}	&	\# investigations involving \{ Parent\} in past \{\# Days\} with \{Allegation\}	\\
\{Parent\}\_INV\_asPERP\_n\_Safety\_Dec\_Unsafe\_\{\# Days\}	&	\# unsafe determinations involving \{ Parent\} in past \{\# Days\}	\\
\{Parent\}\_INV\_asPERP\_n\_Children\_RemHome\_\{\# Days\}	&	\# investigations involving \{ Parent\} in past \{\# Days\} without removal	\\
RECENT\_RPT\_Assigned	&	Child's most recent report was assigned	\\
RECENT\_RPT\_Assigned\_\{Allegation Type\}	&	Child's most recent report was assigned and involved \{Allegation\}	\\
RECENT\_RPT\_CAS\_\{Allegation Type\}	&	Child's most recent report was closed and involved \{Allegation\}	\\
RECENT\_RPT\_Familial	&	Child's most recent report was familial	\\
RECENT\_RPT\_MandatoryReporter	&	Child's most recent report was made by a mandatory reporter	\\
RECENT\_RPT\_Source\_\{Reporter Category\}	&	Child's most recent report was made by \{Reporter Category\}	\\
RECENT\_RPT\_Role\_n\_\{Role\}	&	\# named on child's recent report with a given \{Role\} designation	\\
PA\_CAT	&	Child's protected attribute level	\\
    \bottomrule
    \end{tabular}
\end{table*}

\end{document}